\pdfoutput=1

\documentclass[11pt]{article}

\usepackage{acl}

\usepackage{times}
\usepackage{latexsym}
\usepackage{amssymb}
\usepackage[utf8]{inputenc}
\usepackage{tablefootnote}

\usepackage[T1]{fontenc}

\usepackage{microtype}

\usepackage{times}
\usepackage{latexsym}

\usepackage{graphicx}
\usepackage{amsmath}
\usepackage{algorithmic}
\usepackage{algorithm}
\usepackage{subfloat}
\usepackage{subfig}
\usepackage{microtype}
\usepackage{color}

\newcommand{\etal}{\textit{et al. }}

\usepackage{xurl}
\usepackage{multirow}
\usepackage{microtype}
\usepackage{color}
\usepackage{microtype}
\usepackage{subfloat}

\usepackage[utf8]{inputenc}

\usepackage{sidecap}
\usepackage{times}
\usepackage{latexsym}

\usepackage{graphicx}

\usepackage{amsmath}
\usepackage{algorithmic}
\usepackage{algorithm}
\usepackage{booktabs}

\title{Sparse Progressive Distillation: Resolving  Overfitting under Pretrain-and-Finetune Paradigm}

\author{
    Shaoyi Huang \thanks{~~These authors contributed equally}  \textsuperscript{\rm 1},
    Dongkuan Xu \footnotemark[1]  \textsuperscript{\rm 2},
    Ian En-Hsu Yen \textsuperscript{\rm 3}, \\
    \textbf{Yijue Wang} \textsuperscript{\rm 1},
    \textbf{Sung-En Chang} \textsuperscript{\rm 4},
    \textbf{Bingbing Li} \textsuperscript{\rm 1},
    \textbf{Shiyang Chen} \textsuperscript{\rm 5},\\
    \textbf{Mimi Xie} \textsuperscript{\rm 6},
    \textbf{Sanguthevar Rajasekaran} \textsuperscript{\rm 1},
    \textbf{Hang Liu} \textsuperscript{\rm 5},
    \textbf{Caiwen Ding} \textsuperscript{\rm 1}\\
 $^1$University of Connecticut,
 $^2$Penn State University,
 $^3$Moffett AI,
 $^4$Northeastern University,\\
 $^5$Stevens Institute of Technology,
 $^6$University of Texas at San Antonio\\
 \small\{shaoyi.huang, yijue.wang, sanguthevar.rajasekaran, caiwen.ding\}@uconn.edu,\\
 \small dux19@psu.edu, 
 ian.yan@moffett.ai,
 hliu77@stevens.edu
 }

\begin{document}
\maketitle
\begin{abstract}
Conventional wisdom in pruning Transformer-based language models is that pruning reduces the model expressiveness and thus is more likely to underfit rather than overfit.
However, under the trending pretrain-and-finetune paradigm, we postulate a counter-traditional hypothesis, that is: pruning increases the risk of overfitting when performed at the fine-tuning phase.
In this paper, 
we aim to address the overfitting problem and improve pruning performance via progressive knowledge distillation with error-bound properties. We show for the first time that reducing the risk of overfitting can help the effectiveness of pruning under the pretrain-and-finetune paradigm. Ablation studies and experiments on 
the GLUE benchmark show that our method 
outperforms the leading competitors across different tasks.

\end{abstract}

\section{Introduction}
\label{sec:intro}

Recently, the emergence of Transformer-based language models (using pretrain-and-finetune paradigm)
such as BERT~\cite{devlin2019bert} and GPT-3~\cite{brown2020language} have  revolutionized and established state-of-the-art (SOTA) records (beyond human-level) on various natural language (NLP) processing tasks.
These models are first pre-trained in a self-supervised fashion on a large corpus and fine-tuned for specific downstream tasks~\cite{wang2018glue}. While effective and prevalent, they suffer from redundant computation due to the heavy model size, which hinders their 
popularity on resource-constrained devices, e.g., mobile phones, smart cameras, and autonomous driving~\cite{chen2021re, qi2021accelerating,  yin2021towardsht, yin2021towardsadmm, li2021npas, choi2020edge}. 

\begin{figure}
    \centering
    \subfloat[Pruning under non-pretrain-and-finetune paradigm (e.g., CNN, LSTM, GNN)]{\label{fig:con_prune}
    \hspace{-.0in}\includegraphics[width=.72\columnwidth]{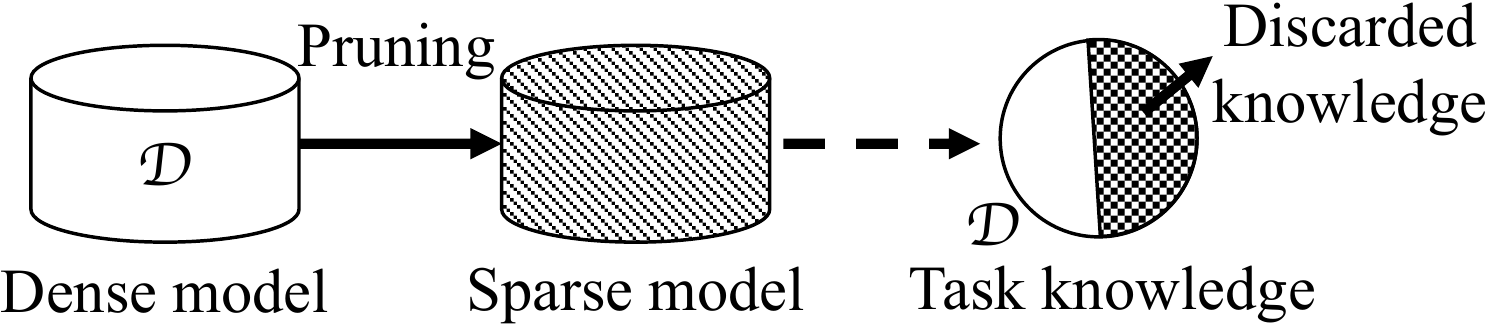}}
    \quad
    \subfloat[Pruning under pretrain-and-finetune paradigm]{\label{fig:finetune_prune}
    \hspace{-.0in}\includegraphics[width=1.\columnwidth]{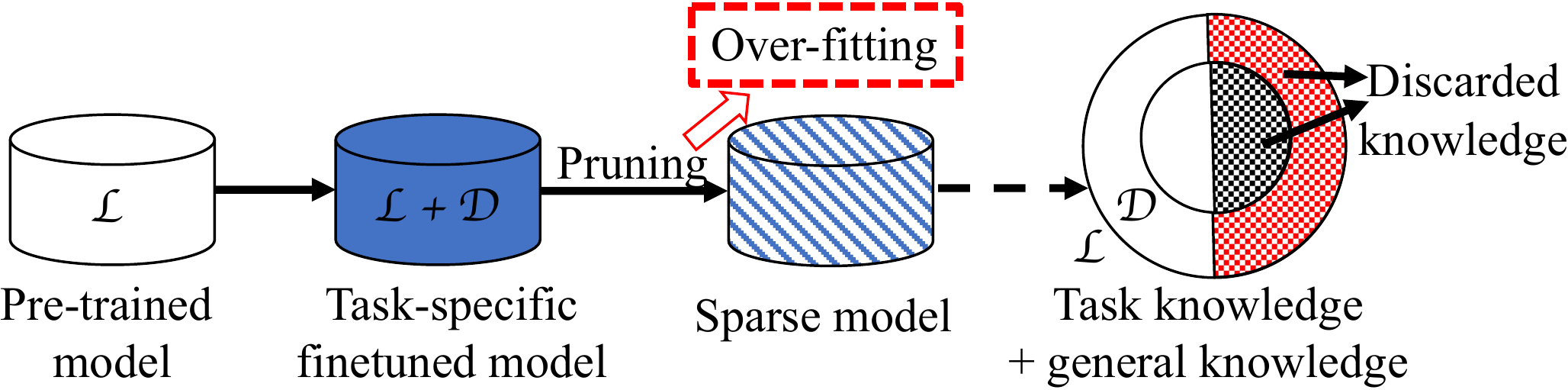}}
    \caption{Pruning under non-pretrain-and-finetune vs. pruning under pretrain-and-finetune. In the subfigures, the cylinders on the left describe the pruning process, and the circles on the right represent the knowledge analysis of the sparse model. 
    }
    \label{fig:motivation-1}
\end{figure}


Various weight pruning approaches (zeroing out certain weights and then optimizing the rest) have been proposed to reduce the footprint requirements of Transformers~\cite{zhu2018prune, blalock2020state,gordon2020compressing,xu2021rethinking,huang2021hmc,peng2021accelerating}.
Conventional wisdom in pruning states that pruning reduces the overfitting risk
since the compressed model structures are less complex, have fewer parameters and are believed to be less prone to overfit~\cite{ying2019overview,wangagainst, tian2020meta, gerum2020sparsity}. However, under the pretrain-and-finetune paradigm, most pruning methods understate the overfitting problem. 

\begin{figure*}
    \centering
    \subfloat[Sparsity=0]{
    \hspace{-.0in}\includegraphics[width=.3\textwidth]{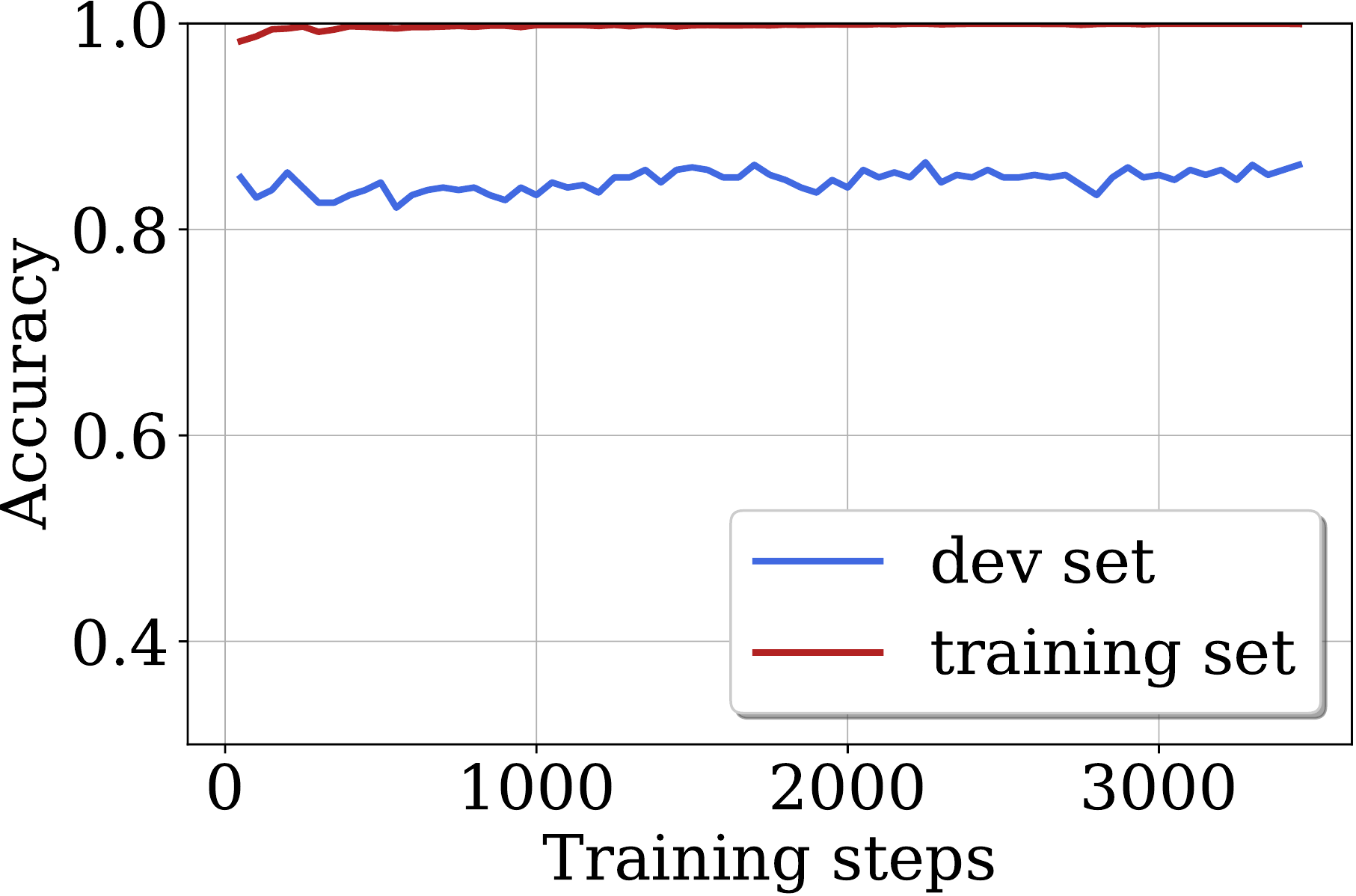}}
    \quad
    \subfloat[Sparsity=0.8]{
    \hspace{-.0in}\includegraphics[width=.3\textwidth]{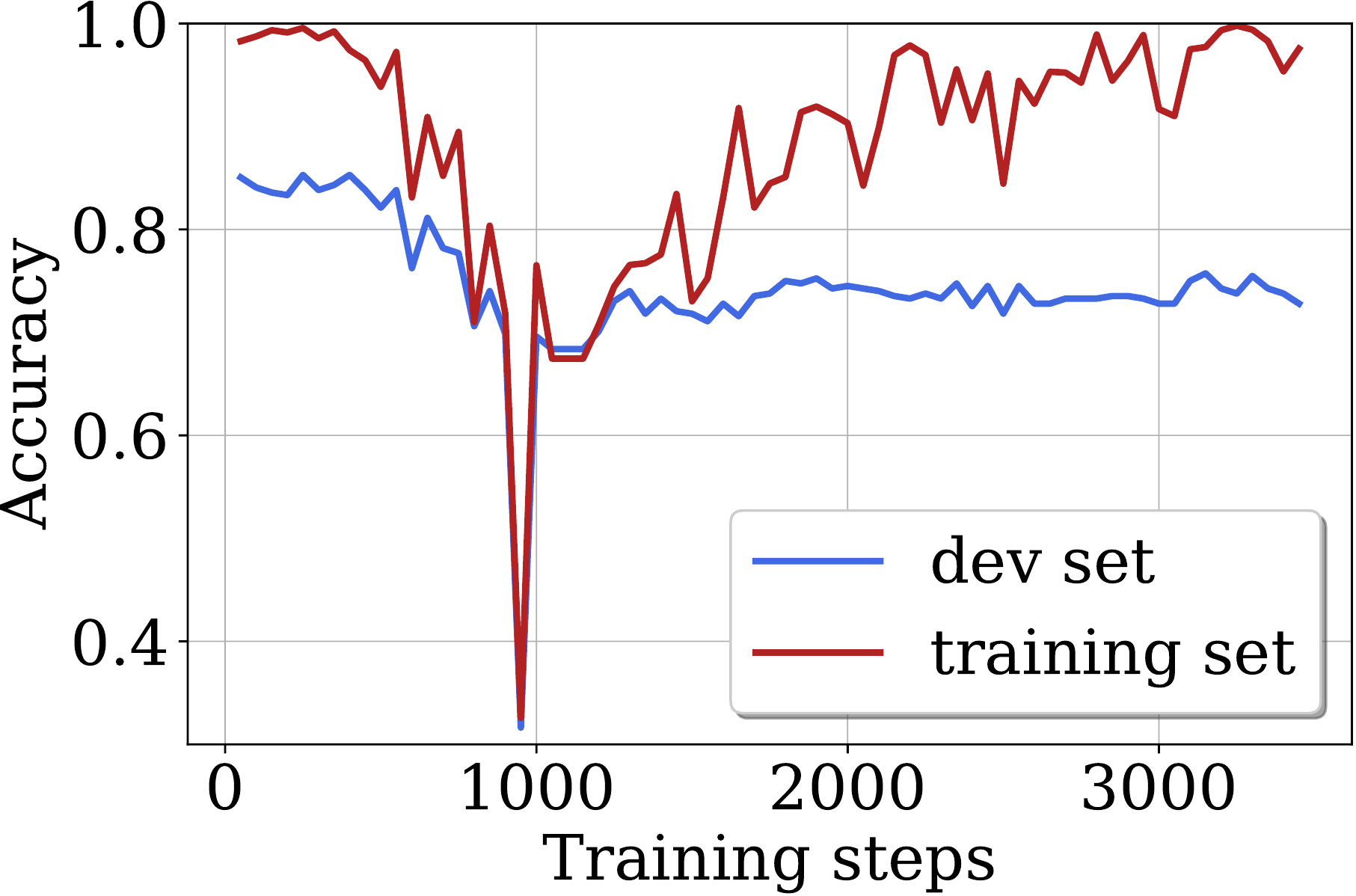}}
    \quad
    \subfloat[Sparsity=0.95]{
    \hspace{-.0in}\includegraphics[width=.3\textwidth]{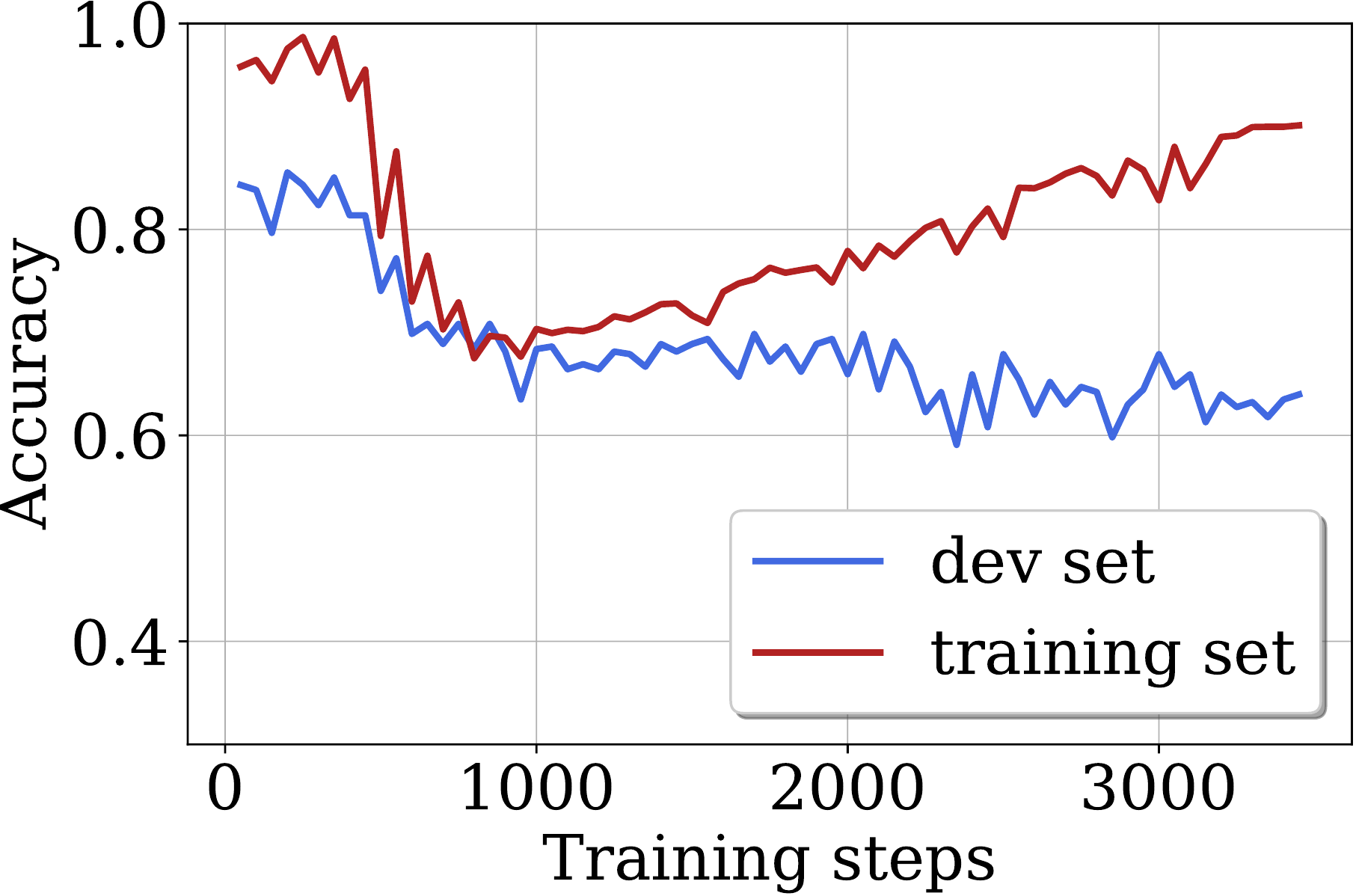}}
    \caption{Visualization of the overfitting problem when pruning weight matrices of BERT$_{\mathrm{BASE}}$ on MRPC at the fine-tuning phase. The overfitting problem becomes more severe with the increasing of sparsity.
    }
    \label{fig:motivation-2}
\end{figure*}

In this paper, we postulate a counter-traditional hypothesis, that is: model pruning increases the risk of overfitting if pruning is performed at the fine-tuning phase.
As shown in Figure~\ref{fig:finetune_prune}, the pretrain-and-finetune paradigm contains two types of knowledge, the general-purpose language knowledge learned during pre-training ($\mathcal{L}$) and the task-specific knowledge from the downstream task data ($\mathcal{D}$). Compared to conventional pruning that only discards task-specific knowledge (Figure~\ref{fig:con_prune}), pruning under pretrain-and-finetune (Figure~\ref{fig:finetune_prune}) discards extra knowledge (red area) learned in pre-training phase. Thus, to recover both the extra discarded general-purpose knowledge and the discarded task-specific knowledge, 
pruning under pretrain-and-finetune increases the amount of information a model needs, which results in relative data deficiency, leading to a higher risk of overfitting. To empirically verify the overfitting problem, we visualize the training and evaluation performance on a real-world task data of MRPC~\cite{devlin2019bert} in Figure~\ref{fig:motivation-2}. From Figure~\ref{fig:motivation-2} (b), it is observed that the evaluation accuracy on the training dataset remains improved while it keeps the same for the validation set through the training process. From Figure~\ref{fig:motivation-2} (c), the difference in performance becomes more significant when the pruning rate becomes higher and the performance on the validation set even becomes worse after 2,000 training steps. All these observations verify our hypothesis.


The main question this paper attempts to answer is: how to reduce the risk of overfitting of pre-trained language models
caused by pruning?
However, answering this question is challenging. First, 
under the pretrain-and-finetune paradigm, 
both the general-purpose language knowledge and the task-specific knowledge are learned. 
It is nontrivial to keep the model parameters related to both knowledge when pruning. 
Second, the amount of data for downstream tasks can be small, 
such as the data with privacy. Thus, the overfitting problem can easily arise, especially in the face of high pruning rate requirements. A little recent progress has been made on addressing overfitting associated with model compression. However, their results are not remarkable and most of them focus on the vision domain~\cite{bai2020few, shen2021progressive}.

To address these challenges, we propose SPD, a \underline{s}parse \underline{p}rogressive \underline{d}istillation method, for pruning pre-trained language models.
\begin{figure*}[h]
    \centering
    \includegraphics[width=1.0\textwidth, height=.35\textwidth]{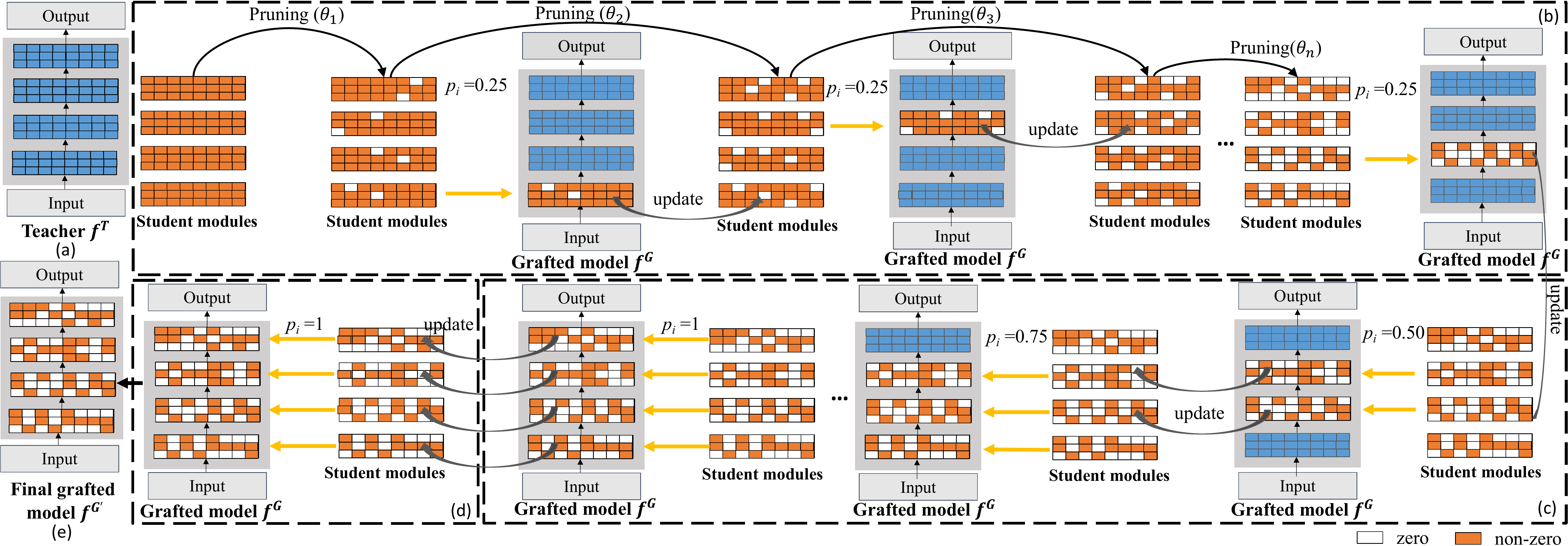}
    \caption{An overview of our sparse progressive distillation method. (a) Teacher model. (b) Pruning to target sparsity. (c) Module grafting with increasing probability. (d) Fine-tuning. (e) Final grafted model.
    }
    \label{fig:main}
\end{figure*} 
We prune and optimize the weight duplicates of the backbone of the teacher model (a.k.a., student modules). 
Each student module shares the same architecture (e.g., the number of weights, the dimension of each weight) as the duplicate.
We replace the corresponding layer(s) of the duplicated teacher model with the pruned sparse student module(s) in a progressive way and name the new model as a grafted model.
We validate our proposed method through the ablation studies and the GLUE benchmark. Experimental results show that our method outperforms the existing approaches. 

We summarize our contributions as follows:
\leftmargini=4mm
\begin{itemize} 
\item We postulate, analyze, and empirically verify a counter-traditional hypothesis: pruning increases the risk of overfitting under the pretrain-and-finetune paradigm.
\item  We propose a sparse progressive pruning method and show for the first time that reducing the risk of overfitting can help the effectiveness of pruning.



\item Moreover, we theoretically analyze that our pruning method can obtain a sub-network from the student model that has similar accuracy as the teacher.

\item Last but not least, we study and minimize the interference between different hyperparameter strategies, including pruning rate, learning rate, and grafting probability, to further improve performance.

\end{itemize}

\section{Related Work}
To summarize, our contribution is determining the overfitting problem of pruning under the pretrain-and-finetune paradigm and proposing the sparse progressive distillation method to address it. We demonstrate the benefits of the proposed framework through the ablation studies. We validate our method on eight datasets from the GLUE benchmark. To test if our method is applicable across tasks, we include the tasks of both single sentence and sentence-pair classification. Experimental results show that our method outperforms the leading competitors by a large margin.  




\noindent\textbf{Network Pruning.}
Common wisdom has shown that weight parameters of deep learning models can be reduced without sacrificing accuracy loss, such as magnitude-based pruning
and lottery ticket hypothesis~\cite{frankle2019lottery}.
\cite{zhu2018prune} compared small-dense models and large-sparse models with the same parameters and showed that the latter outperforms the former,
showing the large-sparse models have better expressive power than their small-dense counterparts. However, under the  pretrain-and-finetune paradigm, pruning leads to overfitting as discussed.

\noindent\textbf{Knowledge Distillation (KD).} 
As a common method in reducing the number of parameters, the main idea of KD is that the small student model mimics the behaviour of the large teacher model and achieves a comparable performance~\cite{hinton2015distilling,mirzadeh2020improved}.
\cite{sanh2019distilbert,jiao2020tinybert,sun2020mobilebert}
utilized KD  to learn universal language representations from large corpus.
However, current SOTA knowledge distillation methods are not able to achieve a high model compression rate (less than 10\% remaining weights) while achieving an insignificant performance decrease.

\noindent\textbf{Progressive Learning.}
The key idea of progressive learning 
is that student learns to update module by module with the teacher. ~\cite{shen2021progressive} utilized a dual-stage distillation scheme where student modules are progressively grafted onto the teacher network, it targets the few-shot scenario and uses only a few unlabeled samples to achieve comparable results on CIFAR-10 and CIFAR-100. ~\cite{xu-etal-2020-bert} gradually increased the probability of replacing each teacher module with their corresponding student module and trained the student to reproduce the behavior of the teacher. 
However, the performance on Transformer-based models of the aforementioned first method is unknown while the second method has an obvious performance drop with a low sparsity (50\%).




\section{Methodology}

\subsection{Problem Formulation}

\begin{figure}
    \centering
    \includegraphics[width=1.0\columnwidth]{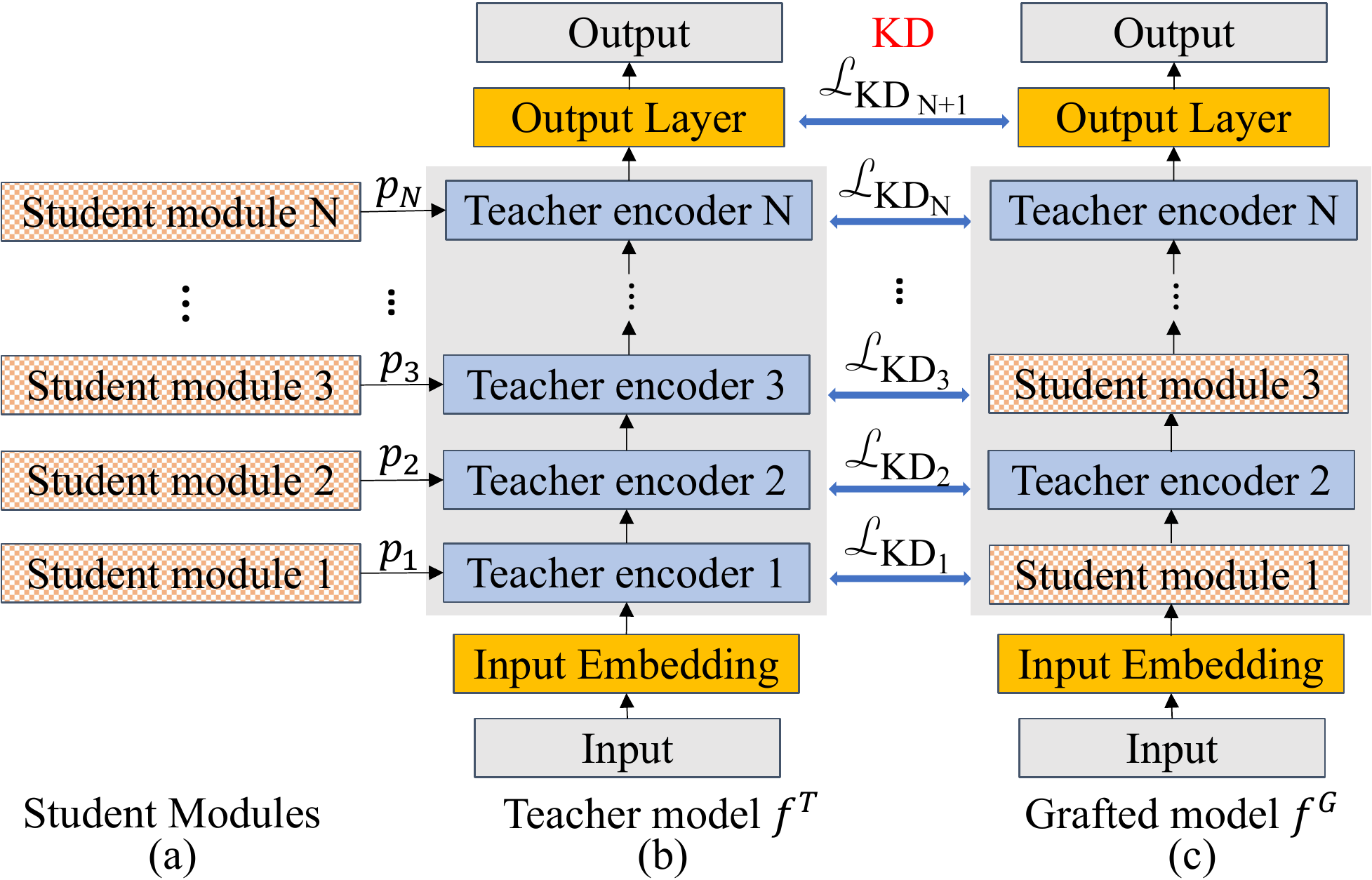}
    \caption{An overview of the layer-wise KD in SPD. {(a) N sparse student modules have probabilities of $p_1$, $p_2$, $p_3$, ..., $p_N$ to substitute the corresponding teacher layers separately. (b) Teacher model. (c) Grafted model. $L_{\text{KD}_i}$ denotes the distillation loss between the $i$-th layer of the teacher and $i$-th layer of the grafted model.}
    }
    \label{fig:distill}
\end{figure}

The teacher model and the grafted model (shown in Figure~\ref{fig:main}) are denoted as {$f^S$} and {$f^G$}, respectively. Both models have $N+1$ layers (i.e., the first $N$ layers are encoder layers, and the $(N+1)$-th layer is the output layer). 
Denote ${{f}^{\mathrm{T}}_{i}(\cdot)}$, ${{f}^{\mathrm{G}}_{i}(\cdot)}$ as the behaviour function induced from the $i$-th encoder of the teacher model, and the grafted model, respectively. As shown in Figure~\ref{fig:distill}, we utilize layer-wise knowledge distillation (KD), where we aim to bridge the gap between ${{f}^{\mathrm{T}}_{i}(\cdot)}$ and ${{f}^{\mathrm{G}}_{i}(\cdot)}$. 


The grafted model is trained to mimic the behavior of the teacher model. 
During training, we minimize the summation loss $\mathcal{L}$:

\begin{equation}
\begin{aligned}
\mathcal{L} = \sum_{{\textbf{x}\in \mathcal{X}}} \sum_{{i=1}}^{N+1} \lambda_i \mathcal{L}_{\text{KD}}({f}^{{T}}_{{i}}({\textbf{x}})
{f}^{{G}}_{{i}}({\textbf{x}})), 
\label{eq:kd} 
\end{aligned}
\end{equation}

where 
$\mathcal{X}$ denotes the training dataset, 
$\lambda_i$ is coefficient of $i$-th layer loss, $\mathcal{L}_{\mathcal{D}}$ is the distillation loss of the layer pair, $x_i$ is the input of the $i$-th layer.

During KD, each student module mimics the behavior of the corresponding teacher layer. Similar to~\cite{jiao2020tinybert}, we take the advantage of abundant knowledge in self-attention distribution, hidden states of each Transformer layer, and the final output layer's soft logits of teacher model to help train the student model. Specifically, we design the KD loss as follows
\begin{equation}
\label{eq:rp}
    \mathcal{L}_{\text{KD}} =
\begin{cases}
\mathcal{L}_{hidn} + \mathcal{L}_{attn} & 1 \leq i \leq N \\
\mathcal{L}_{pred} &  i = N + 1
\end{cases}
\end{equation}

where $\mathcal{L}_{hidn}$ = MSE(${H}^{T}_{i}$, ${H}^{S}_{i}$) ($1 \leq i \leq N$) indicates the difference between hidden states, $\mathcal{L}_{attn}$ = MSE(${A}^{T}_{i}$, ${A}^{S}_{i}$) indicates the difference between attention matrices. 
MSE($\cdot$) is the mean square error loss function and $i$ is the index of Transformer layer. $\mathcal{L}_{pred}$ = -$\text{softmax}$(${z}^{T}$) $\cdot$ $\log\_\text{softmax}$(${z}^{S}$ / $temp$) indicates the difference of soft cross-entropy loss, where ${z}^{T}$ and ${z}^{S}$ are the soft logits of teacher and student model, respectively. $T$ is the temperature hyper-parameter.

We further reduce the number of non-zero parameters in the weight matrix while maintaining accuracy. We denote $\{\textbf{W}_j\}_{j=1}^{j=i}$ as the collection of weights in the first $i$ layers, $\theta_j$ as the sparsity of the $j$-th layer. Then, the loss function of sparse knowledge distillation becomes



\begin{equation}
\small
\begin{aligned}
\mathcal{L} &= \sum_{{\textbf{x}\in \mathcal{X}}}\sum_{{i=1}}^{N+1} \lambda_i \mathcal{L}_{\text{KD}}({f}^{{T}}_{{i}}(\textbf{x},\{\textbf{W}_j\}_{j=1}^{j=i}), {f}^{{G}}_{{i}}(\textbf{x},\{\textbf{W}_j\}_{j=1}^{j=i})) \\ 
& s.t.~~  sparsity(\textbf{W}_j)  \le \theta_j ~ \text{for} ~ j = 1, ..., N \label{eq:sparselossfunction} 
\end{aligned}
\end{equation}


After training, we find the sparse weight matrix $W_{j}^{*}$ using
\begin{equation}
  \textbf{W}_{j}^{*} = {{{\Pi}}_{{{S}}_{j}}}({{\textbf{W}}}_{j})~ \text{for} ~ j = 1, ..., N,
\end{equation}
where ${{{\Pi}}_{{{S}}_{j}}(\cdot)}$ denotes the Euclidean projection onto the set ${{S}}_{j}=\{\textbf{W}_j \mid sparsity(\textbf{W}_j)  \le \theta_j \}.$

\subsection{Our Methods}


\subsubsection{Error-bound Analysis}

Our pruning method is similar to finding matching subnetworks using the lottery ticket hypothesis~\cite{frankle2019lottery,pensia2020optimal} methodology. We analyze the  self-attention (excluding activation). Some non-linear activation functions have been analyzed in~\cite{pensia2020optimal}.

\noindent\textbf{Feed-forward layer.} Consider a feed-forward network $f(x) =w \cdot x$ , and $g(x)=  \left (  \sum_{i=1}^{n} w_i\right )x$.
Lueker \etal\cite{lueker1998exponentially} and Pensia \etal\cite{pensia2020optimal} show that existing a subset of $w_i$, such that the corresponding value of $g(x)$ is very close to $f(x)$.\\
{\bf\em Corollary:} {When $w^*_1,..., w^*_n$ belongs to i.i.d. uniform distribution over [-1,1], where $n \geq C\log\frac{2}{\delta}$, $\delta \leq \text{min}\{1,\epsilon\}$. Then, with probability at least 1-$\delta$, we have 


\begin{equation}
\begin{aligned}
    \exists G_{\textbf{spd}} \subset \{&1,2,...,n\}, \forall w \in [-0.5,0.5],\\
    &s.t \left | w-\sum_{i \in G_{\textbf{spd}}} w^*_i \right | \leq \epsilon 
\end{aligned}
\label{theorem}
\end{equation}
}

\noindent\textbf {Analysis on self-attention.}
The self-attention can be presented as:
\begin{equation}
\begin{aligned}
   \textbf{Z} = \text{attention(\textbf{Q}, \textbf{K}, \textbf{V})}= \text{softmax}(\frac{\textbf{Q}\cdot\textbf{K}^T}{\sqrt{d_k}})\cdot \textbf{V}.
\end{aligned}
\end{equation}
Consider a model $f(x)$ with only one self-attention,
when the token size of input $x$ is 1, $\text{softmax}(\frac{\textbf{Q}\cdot\textbf{K}^T}{\sqrt{d_k}})=1$,
we have \textbf{Z} = \textbf{V}, where $\textbf{V} = w^\textbf{V}x.$

  



Consider $f^G(x)=  \left (  \sum_{i=1}^{d}w_i^G\right )x$
and a pruning sparsity $\mathbf{\theta}$,
base on \textbf{\em{Corollary}}, when $d \geq C\log4/\epsilon$, there exists a pattern of $w_{i}^G$, such that, with probability $1-\epsilon$, 

\begin{equation}
\begin{aligned}
    &\forall w \in [-1,1],  \exists    \theta_i \in \{0,1\},\\
    & s.t.  \left | w- (\sum_{i\in [1,d]} w_{i}^G\mathbb{I}( \theta_i)) \right | < \epsilon
    \label{wplus}
\end{aligned}
\end{equation}
where $\mathbb{I}(\theta_i)$ is the indicator to determine whether $w_{i}^G$ will be remained.

In general, let the token $x$'s size  be $n$. so  $\mathbf{x}=(x_1,x_2,...,x_n)$. Consider a teacher model $f^T(x)$ with a self-attention,
then
\begin{equation}
 \begin{aligned}
   f^T(x_i)&=\text{softmax}(\frac{\textbf{Q}\cdot\textbf{K}^T}{\sqrt{(d_k)}})\cdot\textbf{V}_i \\
   &=(\frac{\sum_{j}e^{c_{ij}}}{\sum _{i}\sum_{j} (e^{c_{ij}})})\cdot\textbf{V}_i \\
   &=(\frac{\sum_{j}e^{c_{ij}}}{\sum _{i}\sum_{j} (e^{c_{ij}})})w^{\textbf{V}_i}x_i\\
   &= w^{c_{i.}} x_i
 \end{aligned}
\end{equation}

where $c_{ij}$ is the $(i,j)^{th}$ element of the matrix $\frac{\textbf{Q}\cdot\textbf{K}^T}{\sqrt{(d_k)}}$.

Base on \textbf{\em{Corollary}}, when $d \geq C\log4/\epsilon$, there exists a pattern of $w_{i}^G$, such that, with probability $1-\epsilon$, 

\begin{equation}
\begin{aligned}
    \forall w^{c_{i.}}\in [-1,1],  \exists  \theta_k \in \{0,1\},\\
    s.t.  \left | w^{c_{i.}}- (\sum_{k \in [1,d]} w_{k}^G\mathbb{I}(\theta_k)) \right |<\epsilon
    \label{wplus}
\end{aligned}
\end{equation}

In summary:
\begin{equation}
\begin{aligned}
    \forall i \in \{1,2,...,n\}, \left | f^T(x_i) - f^G(x_i)\right |<\epsilon
\end{aligned}
\end{equation}

\subsubsection{Progressive Module Grafting}
To avoid overfitting in the training process for the sparse Transformer model, we further graft student modules (scion) onto the teacher model duplicates (rootstock). For the $i$-th student module, we use an independent Bernoulli random variable $\mathbb{I}(\theta_i)$ to indicate whether it will be grafted on the rootstock. To be more specific, $\mathbb{I}(\theta_i)$ has a probability of $p$ (grafting probability) to be set as 1 (i.e., student module substitutes the corresponding teacher layer). Otherwise, the latter will keep weight matrices unchanged. Once the target pruning rate is achieved, we apply linear increasing probability to graft student modules which enable the student modules to orchestrate with each other.




Different from the model compression methods that update all model parameters at once, such as TinyBERT~\cite{jiao2020tinybert} and DistilBERT~\cite{sanh2019distilbert}, SPD only updates the student modules on the grafted model. It reduces the complexity of network optimization, which mitigates the overfitting problem and enables the student modules to learn deeper knowledge from the teacher model. The overview is described in Algorithm~\ref{alg:spd}. We will further demonstrate the effectiveness of progressive student module grafting in~\ref{sec:experimental_results}.

\begin{algorithm}[!h]
\footnotesize
  \caption{Sparse Progressive Distillation}
  \label{alg:spd}
\begin{algorithmic}

  \STATE {\bfseries Input:} Teacher model $f^T$ (fine-tuned BERT$_{\mathrm{BASE}})$; grafted model $f^G$: duplicates of teacher model.
  \STATE {\bfseries Set} $t_1$, $t_2$, $t_3$ as the final number of training steps of pruning, progressive module grafting, and finetuning, respectively.
  \STATE {\bfseries Set} $p$ as the grafting probability
  \STATE {\bfseries Output:} Student model
  
  $p \leftarrow p_0$
  \FOR{t = $0$ to $t_3$ }
        \IF {0 $\leq$ t $< t_1$}
            \STATE Prune student modules and generate mask $M$
            \STATE Graft student modules with $p_0$ 
        \ENDIF
        \IF {$t_1 \leq$ t $ < t_2$}
            \STATE Graft student modules with $p \leftarrow k(t - t_1) + p_0$
        \ENDIF
        \STATE Calculate distillation loss $\mathcal{L}$ in Eqn.~(\ref{eq:sparselossfunction})
        \STATE For $f^G$, update sparse weights \textbf{$w\prime$} $ \leftarrow$ \textbf{$w$} $\cdot$ $M$
        \STATE Duplicate sparse weight(s) on $f^G$ to corresponding student module(s)
  \ENDFOR
    \RETURN $f^G$
\end{algorithmic}
\end{algorithm}



\section{Experiments}

\subsection{Experimental Setup}
\noindent\textbf{Datasets.}
We evaluate SPD on the General Language Understanding Evaluation (GLUE) benchmark~\cite{wang2018glue} and report the metrics, i.e., accuracy scores for SST-2, QNLI, RTE, and WNLI, Matthews Correlation Coefficient (MCC) for CoLA, F1 scores for QQP and MRPC, Spearman correlations for STS-B.



\noindent\textbf{Baselines.} We first use 50\% sparsity (a widely adopted sparsity ratio among SOTA),
and compare SPD against two types of baselines -- non-progressive and progressive. For the former, we select BERT-PKD~\cite{sun2019patient}, DistilBERT~\cite{sanh2019distilbert}, MiniLM~\cite{wang2020minilm}, TinyBERT~\cite{jiao2020tinybert}, SparseBERT~\cite{xu2021rethinking} and E.T.~\cite{chen2021re}, while for the latter, we choose Theseus~\cite{xu-etal-2020-bert}. 
We further compare SPD against other existing works under higher sparsity, e.g., TinyBERT~\cite{jiao2020tinybert}, SparseBERT~\cite{xu2021rethinking} and RPP~\cite{guo2019reweighted}.


\noindent\textbf{SPD Settings.} We use official {BERT$_{\mathrm{BASE}}$}, uncased model as the pre-train model and the fine-tuned pre-train model as our teacher. Both {BERT$_{\mathrm{BASE}}$} and teacher model have the same architecture (i.e., 12 encoder layers (L = 12; embedding dimension $d_{model}$ = 768; self-attention heads H = 12)). We finetune {BERT$_{\mathrm{BASE}}$} using 
best performance from \{$2e^{-5}, 3e^{-5}, 4e^{-5}, 5e^{-5}$\} as the learning rate.
For SPD model training, the number of pruning epochs, linear increasing module grafting epochs, finetuning epochs vary from [10, 30], [5, 20], [5, 10], respectively. For pruning, we use AdamW~\cite{loshchilov2018fixing} as the optimizer and run the experiments with an initial grafting probability from \{0.1, 0.2, 0.3, 0.4, 0.5, 0.6, 0.7, 0.8, 0.9\}. The probability with the best performance will be adopted. After pruning, we adjust the slope of the grafting probability curve so that the grafting probability equals 1 at the end of module grafting. For module grafting and finetuning, an AdamW optimizer is used with learning rate chosen from \{$3e^{-5}$, $1e^{-4}$, $3.2e^{-4}$, $5e^{-4}$, $6.4e^{-4}$\}. The model training and evaluation are performed with
CUDA 11.1 on Quadro RTX6000 GPU and Intel(R) Xeon(R) Gold 6244 @ 3.60GHz CPU.



\subsection{Experimental Results}
\label{sec:experimental_results}

\begin{table*}[!h]
\centering
\resizebox{1\textwidth}{!}{
\begin{tabular}{lcccccccccc}
\toprule
\multirow{3}*{\textbf{Model}} & \multirow{3}*{\textbf{\#Param}} & \textbf{MNLI} & \textbf{QQP} & \textbf{QNLI} & \textbf{SST-2} & \textbf{CoLA} & \textbf{STS-B} &\textbf{MRPC}  & \textbf{RTE} & \multirow{3}*{\textbf{Avg.}}\\
& &(393k) & (364k) & (105k) & (67k) & (8.5k) & (5.7k) & (3.7k) & (2.5k) \\
& & Acc & F1 & Acc & Acc & Mcc & Spea & F1 & Acc \\
\midrule
\textbf{BERT$_{\mathrm{BASE}}$~\cite{devlin2019bert}}      & 109M       & 84.6& 91.2 & 90.5  & 93.5  & 52.1  &  85.8  & 88.9 & 66.4 & 81.6 \\
\textbf{BERT$_{\mathrm{BASE}}$ (ours)}      & 109M       & 83.9 & 91.4 & 91.1  & 92.7  & 53.4  &  85.8  & 89.8 & 66.4 & 81.8 \\
\textbf{Fine-tuned BERT$_{\mathrm{BASE}}$ (teacher)}     & 109M       & 84.0 & 91.4 & 91.6  & 92.9  & 57.9  &  89.1  & 90.2 & 72.2 & 83.7\\
\midrule
\multicolumn{11}{c}{\textbf{\textit{non-progressive}}} \\
\textbf{BERT$_6$-PKD~\cite{sun2019patient}}    & 67M   &81.5 &88.9 & 88.4  & 91.0  & 45.5  &  86.2  & 85.7 & 66.5 & 79.2 \\
\textbf{DistilBERT~\cite{sanh2019distilbert}}      & 67M    &82.2 & 88.5 & 89.2  & 92.7  & 51.3  &  86.9  & 87.5 & 59.9 & 79.8 \\
\textbf{MiniLM$_6$~\cite{wang2020minilm}}     & 67M    & 84.0 & 91.0 & 91.0  & 92.0  & 49.2  &  -  & 88.4 & 71.5 & - \\
\textbf{TinyBERT$_6$~\cite{jiao2020tinybert}}   & 67M    &84.5 &91.1 & 91.1  & 93.0  & 54.0  &  90.1  & 90.6  & \textbf{73.4} & 83.5\\
\textbf{SparseBERT~\cite{xu2021rethinking}} & 67M    & 84.2 & 91.1 & 91.5  & 92.1  & 57.1 & 89.4 & 89.5  & 70.0 & 83.1  \\
\textbf{E.T.~\cite{chen2021re}} & 67M    & 83.7 & 86.5 & 88.9  & 90.8  & 55.6 & 87.6 & 88.7  & 69.5 & 81.4  \\
\midrule
\multicolumn{11}{c}{\textbf{\textit{progressive}}} \\
\textbf{Theseus~\cite{xu-etal-2020-bert}} & 67M    & 82.3 & 89.6 & 89.5  & 91.5  & 51.1 & 88.7 & 89.0  & 68.2 &  81.2\\

\textbf{SPD (ours)} & 67M    &\textbf{85.0} & \textbf{91.4} & \textbf{92.0}  & \textbf{93.0}  & \textbf{61.4} & \textbf{90.1} & \textbf{90.7}  & 72.2 &  \textbf{84.5}\\
\bottomrule
\end{tabular}}
\caption{Results on the dev set of the GLUE benchmark. The results of DistilBERT and TinyBERT$_6$ are taken from ~\cite{jiao2020tinybert}. Mcc refers to Matthews correlation coefficient, and Spea refers to Spearman correlation coefficient.}
\label{tb:compress_2}
\vspace{0.2cm}
\end{table*}

\begin{table}[t]
\footnotesize
\addtolength{\tabcolsep}{-5pt}
\renewcommand{\arraystretch}{1}
\centering
\begin{tabular}{lccccccc}
\toprule
\multirow{2}*{\textbf{Model}} & \multirow{2}*{Sparsity}  & CoLA & STS-B &MRPC  & RTE & \multirow{2}*{\textbf{Avg.}}\\
                     & & (Mcc) & (Spea) & (F1) & (Acc)  \\
\midrule
Teacher       & 100\%         & 57.9  &  89.1  & 90.2 & 72.2 & 77.4\\
\midrule
TinyBERT$_4$$^*$    & 82\%      & 29.8  &  -  & 82.4 & - & -\\
RPP       & 88.4\%     & -  &  -  & 81.9 & 67.5 & -  \\
SparseBERT$^*$       & 95\%     & 18.1  & 32.2  & 81.5  &  47.3 & 44.8 \\
\midrule
SPD (\textbf{ours})   & 66.6\%     & 50.7  &  88.9  & 90.4 & 69.7  & 74.9\\
SPD (\textbf{ours})   & 75\%     & 50.0  &  88.3  & 90.2 & 67.9  & 74.1\\
SPD (\textbf{ours})   & 87.5\%     & 49.9  &  87.8  & 89.9 & 67.9 & 73.9  \\
SPD (\textbf{ours})   & 90\%     & 48.7  &  87.8  & 89.9 & 69.0  & 73.9\\
SPD (\textbf{ours})   & 95\%     & 42.1  &  86.9  & 88.7 & 56.7  & 68.2\\
\bottomrule
\end{tabular}
\caption{Results on the dev set of the GLUE benchmark at higher pruning rates. $^*$ denotes data augmentation removed for the fair comparison.
}
\label{tb:compress_higher}
\vspace{0.02cm}
\end{table}

\noindent\textbf{Accuracy vs. Sparsity.}
We do experiments on eight GLUE benchmark tasks (Table~\ref{tb:compress_2}).
For non-progressive baselines, SPD exceeds all of them on QNLI, SST-2, CoLA, STS-B, and MRPC. For RTE, TinyBERT$_6$ has a 1.6\% higher accuracy than SPD. However, TinyBERT$_6$ used augmented data while SPD does not use data augmentation to generate the results in Table~\ref{tb:compress_2}. On average, SPD has 6.3\%, 5.6\%, 1.2\%, 1.7\%, 3.7\% improvement in performance than BERT$_6$-PKD, DistilBERT, TinyBERT$_6$, SparseBERT, E.T. respectively. Furthermore, on CoLA, SPA achieves up to 25.9\% higher performance compared to all non-progressive baselines. For the progressive baseline, we compare SPD with BERT-of-Theseus. Experimental results show that SPD exceeds the latter on all tasks. SPD has a 3.9\% increase on average. Among all the tasks, CoLA and RTE have 20.2\% and 5.9\% gain respectively.
For the comparison with sparse and non-progressive baseline, SPD has an improvement of 16.8\%, 5.5\%, 3.2\%, 2.7\%, 2.0\%, 1.9\%, 1.6\%, 1.6\% on CoLA, RTE, MNLI, QNLI, QQP, MRPC, STS-B, SST-2, respectively.


On all listed tasks, SPD even outperforms the teacher model except for RTE. On RTE, SPD retains exactly the full accuracy of the teacher model. On average, the proposed SPD achieves a 1.1\% higher accuracy/score than the teacher model. We conclude the reason for the outstanding performance from three respects: 1) There is redundancy in the original dense BERT 
model. Thus, pruning the model with a low pruning rate (e.g., 50\%) will not lead to a significant performance drop. 2) SPD decreases the overfitting risk which helps the student model learn better. 
3) The interference between different hyperparameter strategies is mitigated, which enables SPD to obtain a better student model.




\begin{figure*}[!h]
    \centering
    \subfloat[No progressive, no KD]{
    \hspace{-.05in}\includegraphics[width=.225\textwidth]{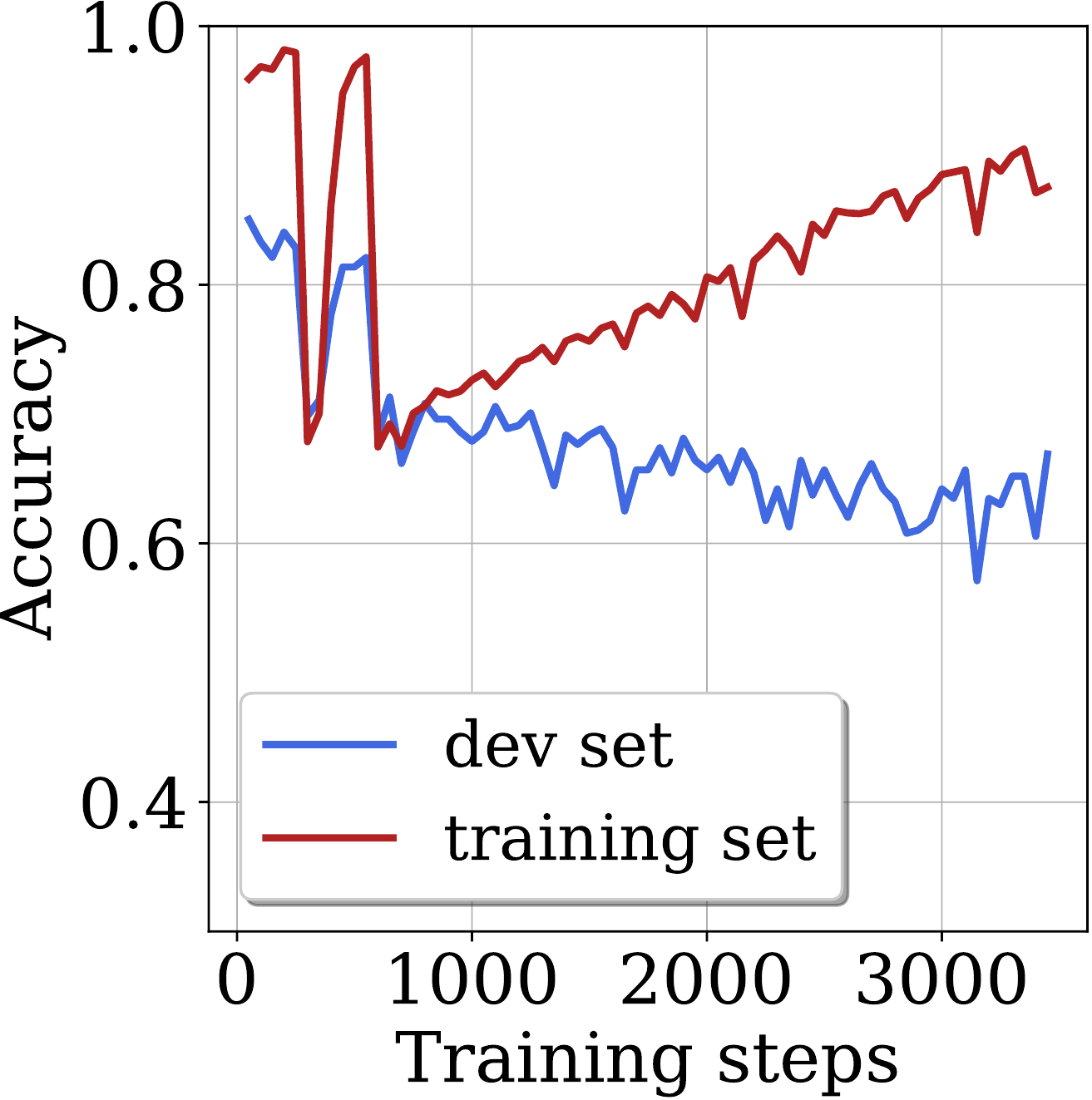}}
    \quad
    \subfloat[Progressive, no KD]{
    \hspace{-.05in}\includegraphics[width=.225\textwidth]{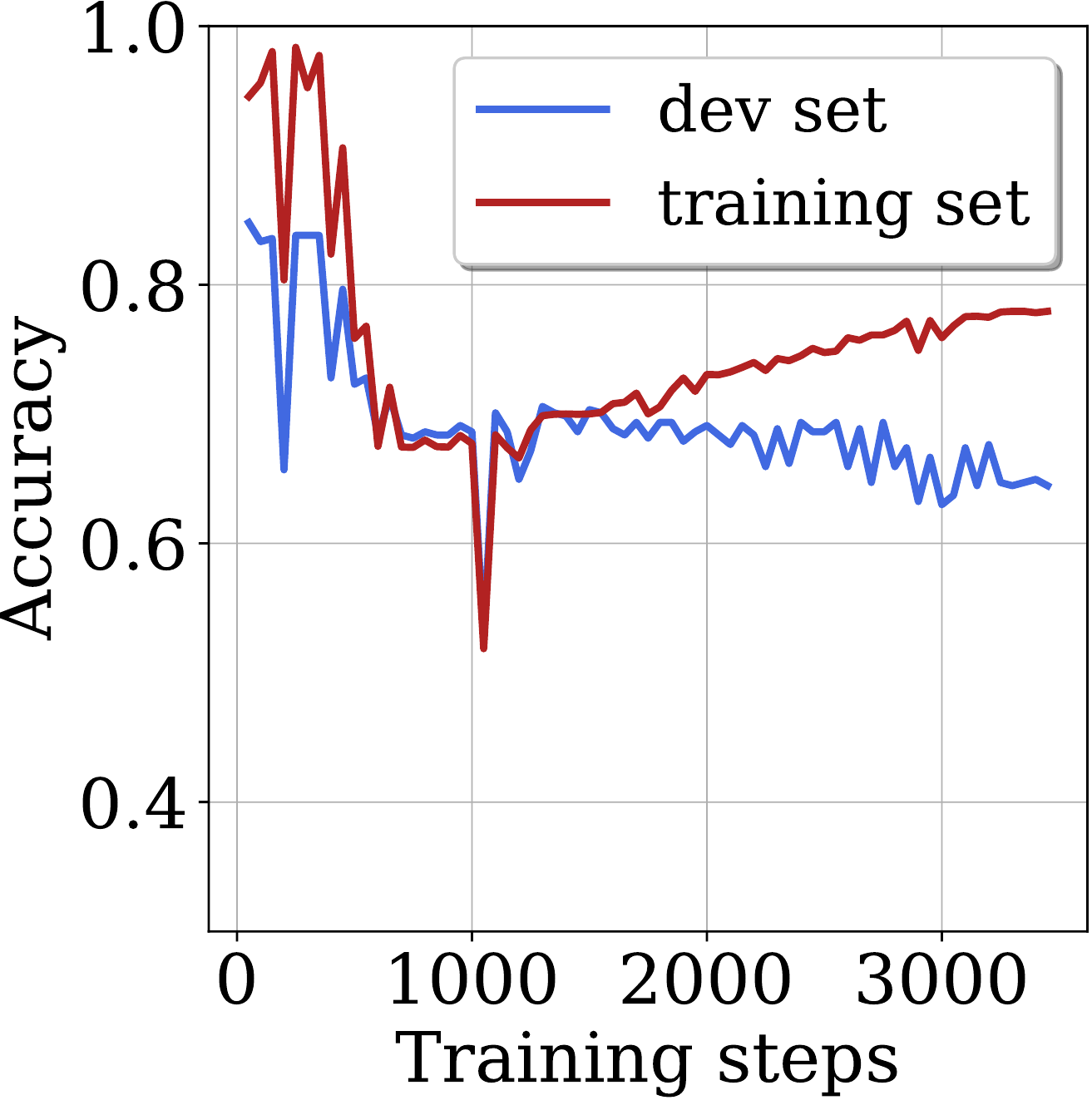}}
    \quad
    \subfloat[No progressive, KD]{
    \hspace{-.05in}\includegraphics[width=.225\textwidth]{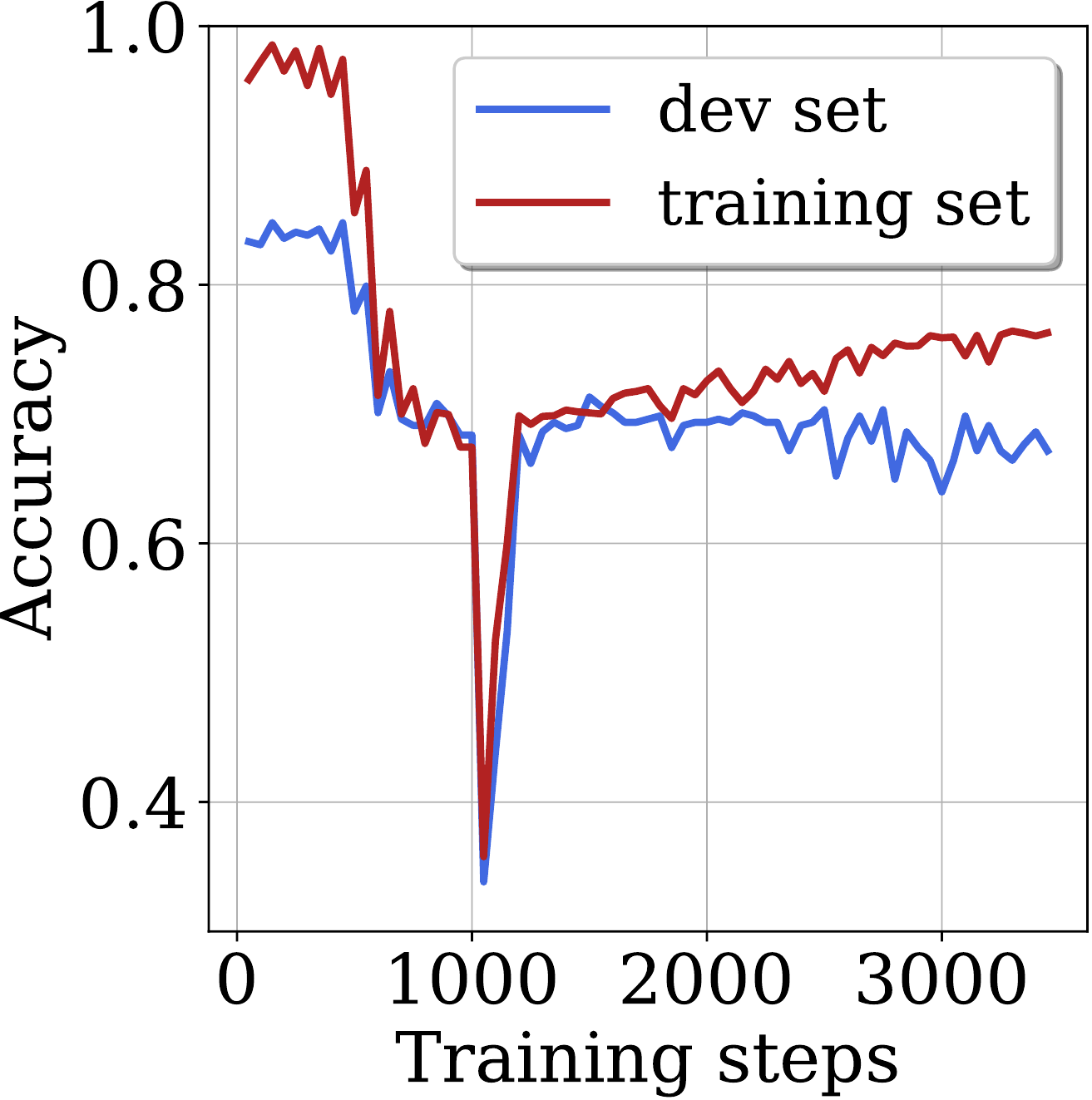}
    }
    \quad
    \subfloat[Progressive, KD (ours)]{
    \hspace{-.05in}\includegraphics[width=.225\textwidth]{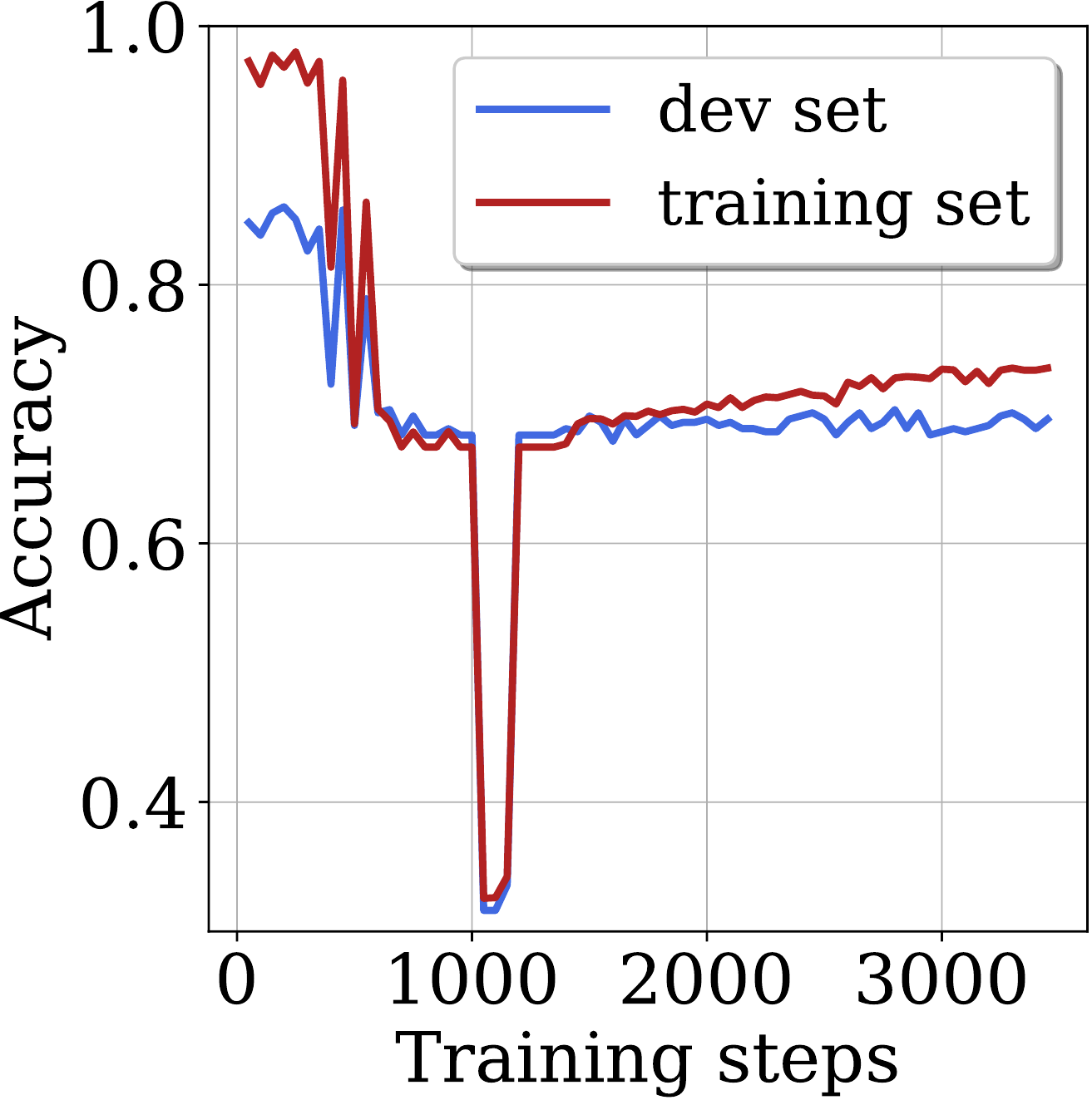}}
    \caption{Comparison of four strategies to deal with the overfitting problem on MRPC.}
    \label{fig:overfitting}
\end{figure*}

We also compare SPD with other baselines (i.e., 4-layer TinyBERT~\cite{jiao2020tinybert}, RPP~\cite{guo2019reweighted}, and SparseBERT~\cite{xu2021rethinking}) under higher pruning rates. Results are summarized in Table~\ref{tb:compress_higher}. For the fairness of comparison, we remove data augmentation from the above methods. 
We mainly compare the aforementioned baselines with very high sparsity (e.g., 90\%, 95\%) SPD. For the comparison with TinyBERT$_4$, both SPD ($90\%$ sparsity) and SPD ($95\%$ sparsity) win. SPD ($90\%$ sparsity) has 63.4\% and $9\%$ higher evaluation score than TinyBERT$_4$ on CoLA and MRPC, respectively. 
For the setting of $95\%$ sparsity, SPD outperforms TinyBERT$_4$ with $41.3\%$ and $7.6\%$ higher performance, respectively. Compared to RPP, both SPD ($90\%$ sparsity) and SPD ($95\%$ sparsity) show higher performance on MRPC, with $9.8\%$ and $8.3\%$ higher F1 score, respectively. For SparseBERT, SPD exceeds it on all tasks in Table~\ref{tb:compress_higher}. Especially on CoLA, SPD ($90\%$ sparsity) and SPD ($95\%$ sparsity) have 2.69$\times$ and 2.33$\times$ higher Mcc score on CoLA, respectively. SparseBERT has competitive performance with SOTA when using data augmentation. The reason for the performance drop for SparseBERT may because its deficiency of ability in mitigating overfitting problems.


\noindent\textbf{Overfitting Mitigation.} We explore the effectiveness of SPD to mitigate the overfitting problem. 
Depending on whether progressive, grafting, or KD is used, we compare 4 strategies: (a) no progressive, no KD; (b) progressive, no KD; (c) no progressive, KD; (d) progressive, KD (ours). We evaluate these strategies on both training and validation sets of MRPC. The results are summarized in Figure~\ref{fig:overfitting}. 
From (a) to (d), the gap between the evaluation results of the training set and the dev set is reduced, which strongly suggests that the strategy adopted by SPD, i.e., progressive + KD, outperforms other strategies in mitigating the overfitting problem.
Figure~\ref{fig:overfitting} (a), (b), and (c) indicate that compared to progressive only, KD has a bigger impact on mitigating overfitting, as 
the performance gap between the training set and the dev set decreases more from (a) to (c) than from (a) to (b). 
From Figure~\ref{fig:overfitting} (a), (b) and (c), we also observe that compared to no progressive, no KD, either using progressive (Figure~\ref{fig:overfitting} (b)) or KD (Figure~\ref{fig:overfitting} (c)) is very obvious to help mitigate the overfitting problem. Figures~\ref{fig:overfitting} (b), (c) and (d) indicate that the combination of progressive and KD brings more benefits than only using progressive or KD as Figure~\ref{fig:overfitting} (d) has the smallest performance gap between the training set and the dev set.
Combined with Table~\ref{tb:compress_2} and Table~\ref{tb:compress_higher}, Figure~\ref{fig:overfitting} shows that SPD mitigates overfitting and leads to higher performance. 

\subsection{Ablation Studies}

\begin{table*}[!h]
\centering
\resizebox{1\textwidth}{!}{
\begin{tabular}{lcccccccccc}
\toprule
\multirow{2}*{\textbf{Model}} & \multirow{2}*{\textbf{\#Param}} & \textbf{MNLI} & \textbf{QQP} & \textbf{QNLI} & \textbf{SST-2} & \textbf{CoLA} & \textbf{STS-B} &\textbf{MRPC}  & \textbf{RTE} & \multirow{2}*{\textbf{Avg.}}\\
& & Acc & F1 & Acc & Acc & Mcc & Spea & F1 & Acc \\
\midrule
\textbf{Fine-tuned BERT$_{\mathrm{BASE}}$ (teacher)}     & 109M       & 84.0 & 91.4 & 91.6  & 92.9  & 57.9  &  89.1  & 90.2 & 72.2 & 83.7 \\
\midrule
\textbf{Sparse + KD} & 67M    & 84.2 & 91.1 & 91.5  & 92.1  & 57.1 & 89.4 & 89.5  & 70.0 & 83.1 \\
\textbf{Sparse + Progressive} & 67M    & 83.9 & 91.2 & 91.5  & 92.3  & 57.4 & 89.6 & 89.6  & 71.4 & 83.4 \\
\textbf{SPD (ours)} & 67M    &\textbf{85.0} & \textbf{91.4} & \textbf{92.0}  & \textbf{93.0}  & \textbf{61.4} & \textbf{90.1} & \textbf{90.7}  & \textbf{72.2} & \textbf{84.5} \\
\bottomrule
\end{tabular}}
\caption{The performance comparison of different strategies on the dev set of GLUE. Mcc refers to Matthews correlation coefficient and Spea refers to Spearman correlation coefficient.
}
\label{tb:component}
\vspace{0.2cm}
\end{table*}

In this section, 
we justify the three schedulers used in our method (i.e., grafting probability, pruning rate, and learning rate), and study the sensitivity of our method with respect to each of them.



\noindent\textbf{Study on Components of SPD.} The proposed SPD consists of three components (i.e., sparse, knowledge distillation, and progressive module grafting). We conduct experiments to study the importance of each component on GLUE benchmark tasks with the sparsity of 50\% and results are shown in Table~\ref{tb:component}. Compared to both sparse + KD and sparse + progressive, SPD achieves gains on performance among all tasks. 

\begin{figure}[!h]
\centering
\includegraphics[width=1.\columnwidth]{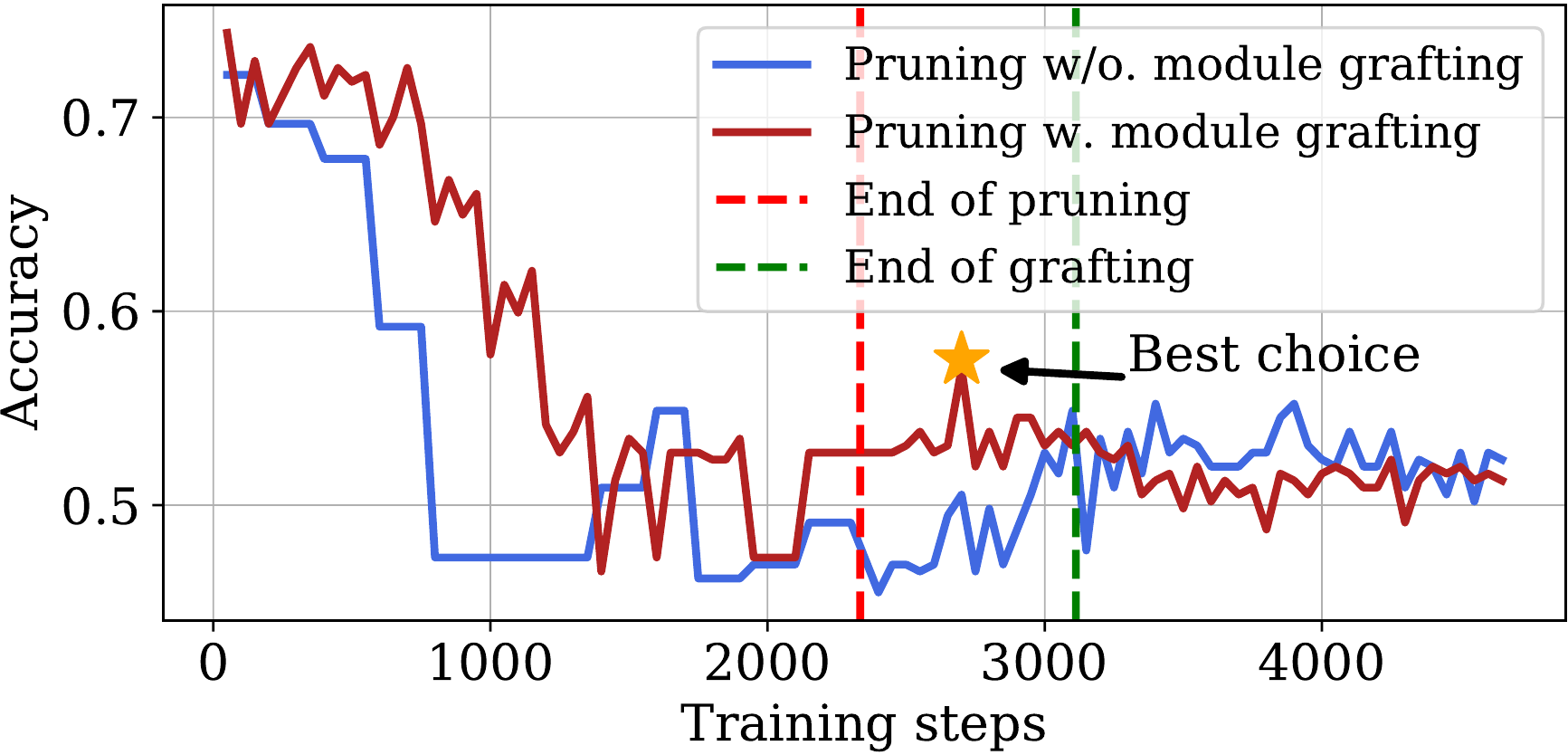}
\caption{Pruning w/ module grafting vs. Pruning w/o. module grafting on RTE (dev set).}
\label{fig:stage_one_with_scheduler_better}
\end{figure}
\begin{figure}[!h]
\centering
\includegraphics[width=1.\columnwidth]{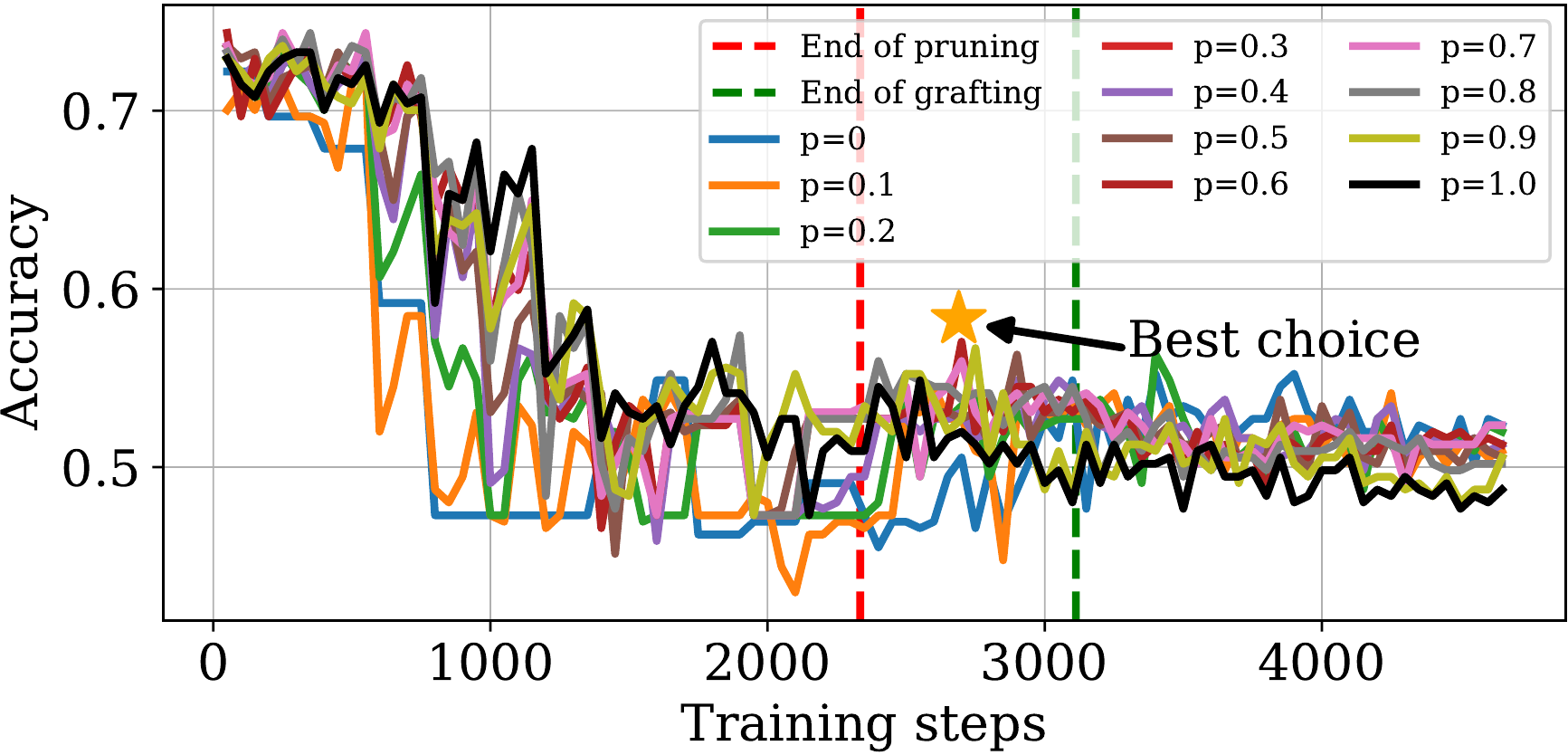}
\caption{Sensitivity analysis of grafting probability on RTE (dev set).}
\label{fig:different_p0}
\end{figure}

\noindent\textbf{Effects of Grafting Probability Strategy.}
In our method, we set the grafting probability greater than 0 during pruning, to allow student modules to learn deeper knowledge from the teacher model.
To verify the benefit of this design, 
we change the grafting probability to zero and compare it with our method. The result on RTE is shown in Figure~\ref{fig:stage_one_with_scheduler_better}. Pruning with grafting (the red curve) shows better performance than pruning without grafting, which justifies the existence of grafting during pruning. In addition, we study the sensitivity of our method to grafting probability (Figure~\ref{fig:different_p0}). It is observed that $p_0$ = 0.6 achieves the best performance,  and the progressive design is better than the non-progressive. 

\vspace{.1in}

\begin{figure}[!h]
\centering
\includegraphics[width=1\columnwidth]{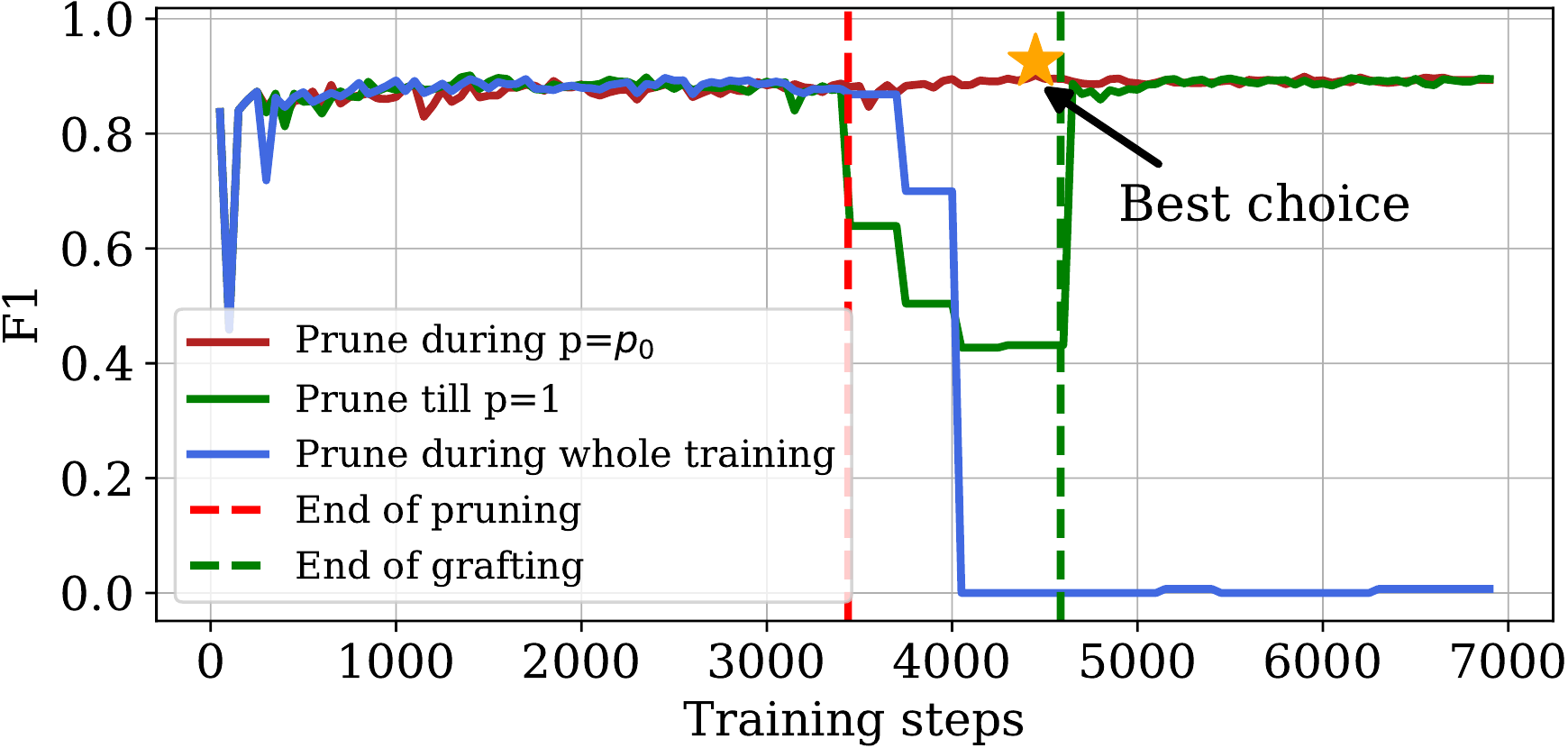}
\caption{Effects of different pruning ending strategies on MRPC (dev set).}
\label{fig:pruning}
\end{figure} 




\noindent\textbf{Effects of Pruning Rate Strategy.}
For the pruning rate scheduler, we compare the strategies with different pruning ending steps. 
The results are shown in Figure~\ref{fig:pruning}. It is observed that the pruning during when grafting probability $p$ = $p_0$ has a higher F1 score than other strategies on MRPC.

\begin{figure}[!h]
    \centering
    \subfloat[One optimizer]{
    \hspace{.2in}\includegraphics[width=.19\textwidth, height=.19\textwidth]{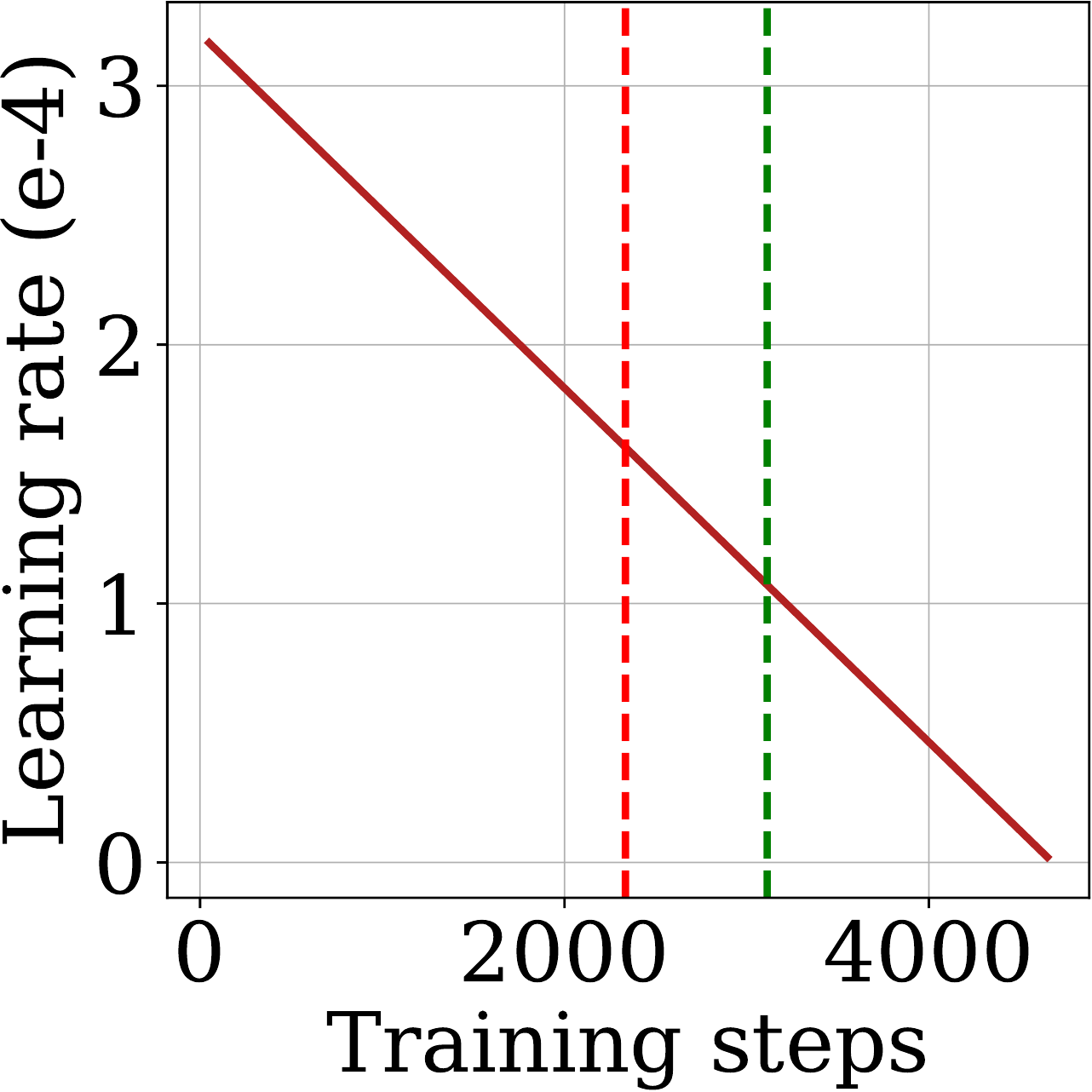}
    }
    \quad
    \subfloat[Two optimizers]{
    \hspace{.1in}\includegraphics[width=.19\textwidth, height=.19\textwidth]{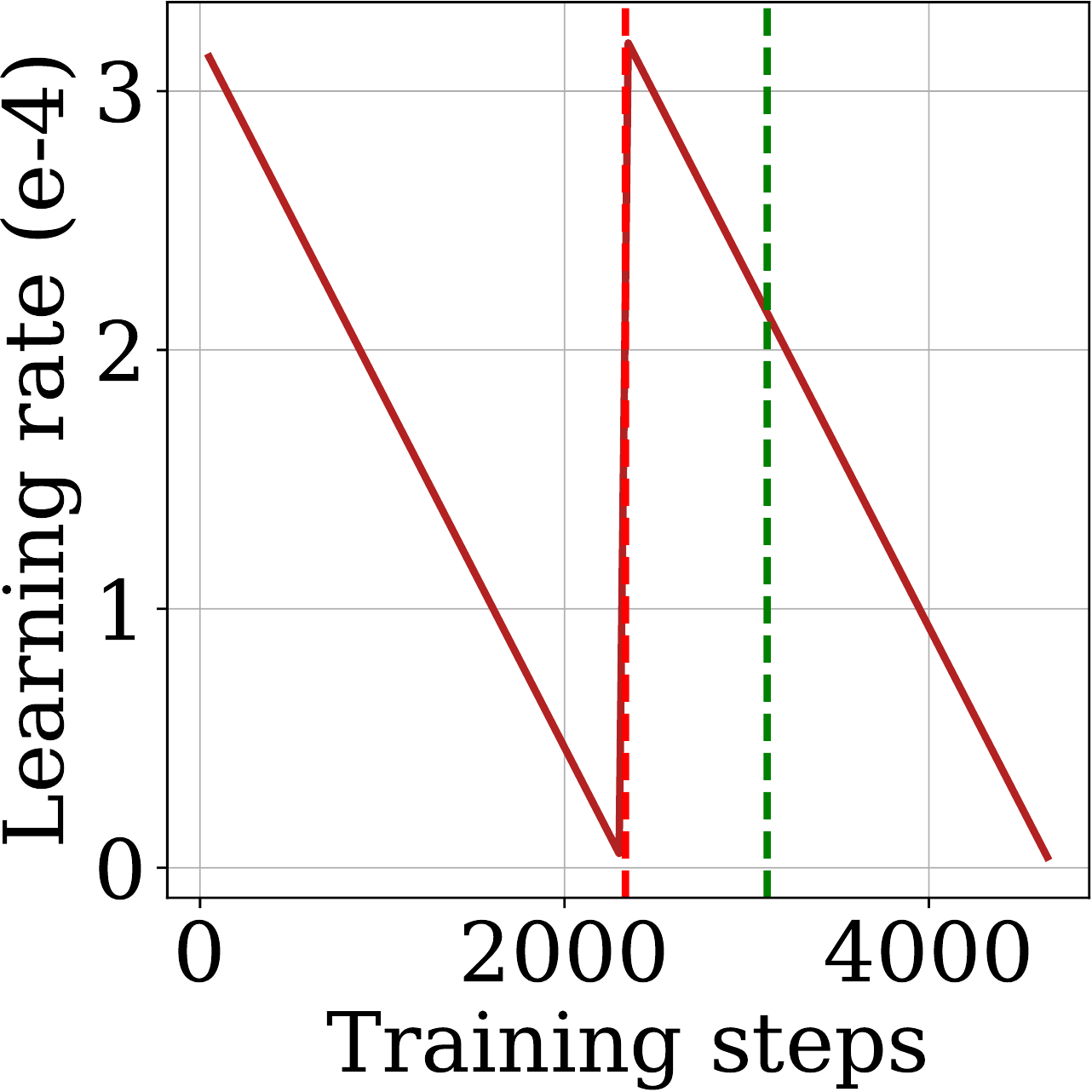}}
    \quad
    \subfloat[Comparison of two different optimizer settings]{
    \hspace{-.0in}\includegraphics[width=1.\columnwidth]{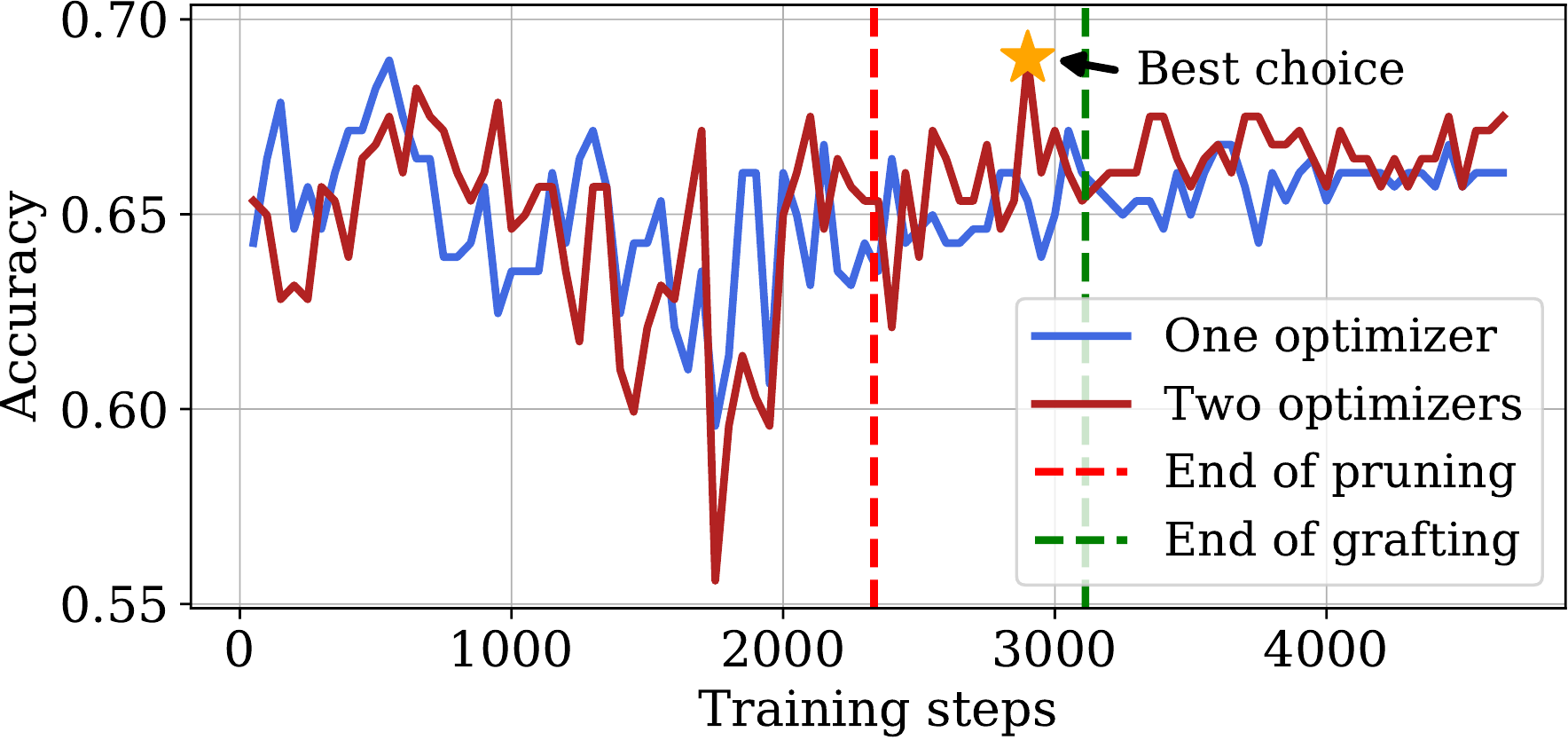}
    }
    \quad
    \caption{(a) The learning rate curve of one AdamW optimizer in training. (b) The learning rate of two AdamW optimizers in training. (c) Performance comparison of the above two settings.}
    \label{fig:optimizer}
\end{figure}

\noindent\textbf{Effects of Optimizer Strategy.}
We also compare our strategy with the strategy that only has one learning rate scheduler. The results 
(Figure~\ref{fig:optimizer}) indicate that our strategy (i.e., two independent optimizers) is better. We also evaluate different learning rates 
with the pruning rate of 0.9 and the grafting probability of 0.8. 

\section{ Conclusion}
In this paper, we postulate a counter-traditional hypothesis that pruning increases the risk of overfitting under the pretrain-and-finetune paradigm. We analyze and empirically verify this hypothesis, and propose a sparse progressive pruning method to address the overfitting problem. 
We theoretically analyze that our pruning method can obtain a sub-network from the student model that has a similar accuracy as the teacher.
We study and minimize the interference between different hyperparameter strategies, including pruning rate, learning rate, and grafting probability.
A number of ablation studies and experimental results on eight tasks from the GLUE benchmark demonstrate the superiority of our method over the leading competitors.


\section*{Acknowledgement}
This research was supported in part by National Science Foundation (NSF) CRII Award No. 2000722 and NSF CAREER Award No. 2046102. Sanguthevar Rajasekaran has been supported in part by the NSF RAISE Award No. 1743418 and NSF EAGER Award No. 1843025. In addition, it used the Extreme Science and Engineering Discovery Environment (XSEDE) through allocations TG-CCR200004.

\bibliography{custom}
\bibliographystyle{acl_natbib}

\clearpage

\section*{Appendix}

We provide the sensitivity analysis of learning rate on RTE and STS-B (dev set) and the evaluation curves of four tasks (CoLA, STS-B, MRPC, and RTE) with the target pruning rate of 0.95.

\begin{figure}[h]
\centering
\vspace{.14in}
\includegraphics[width=1.\columnwidth]{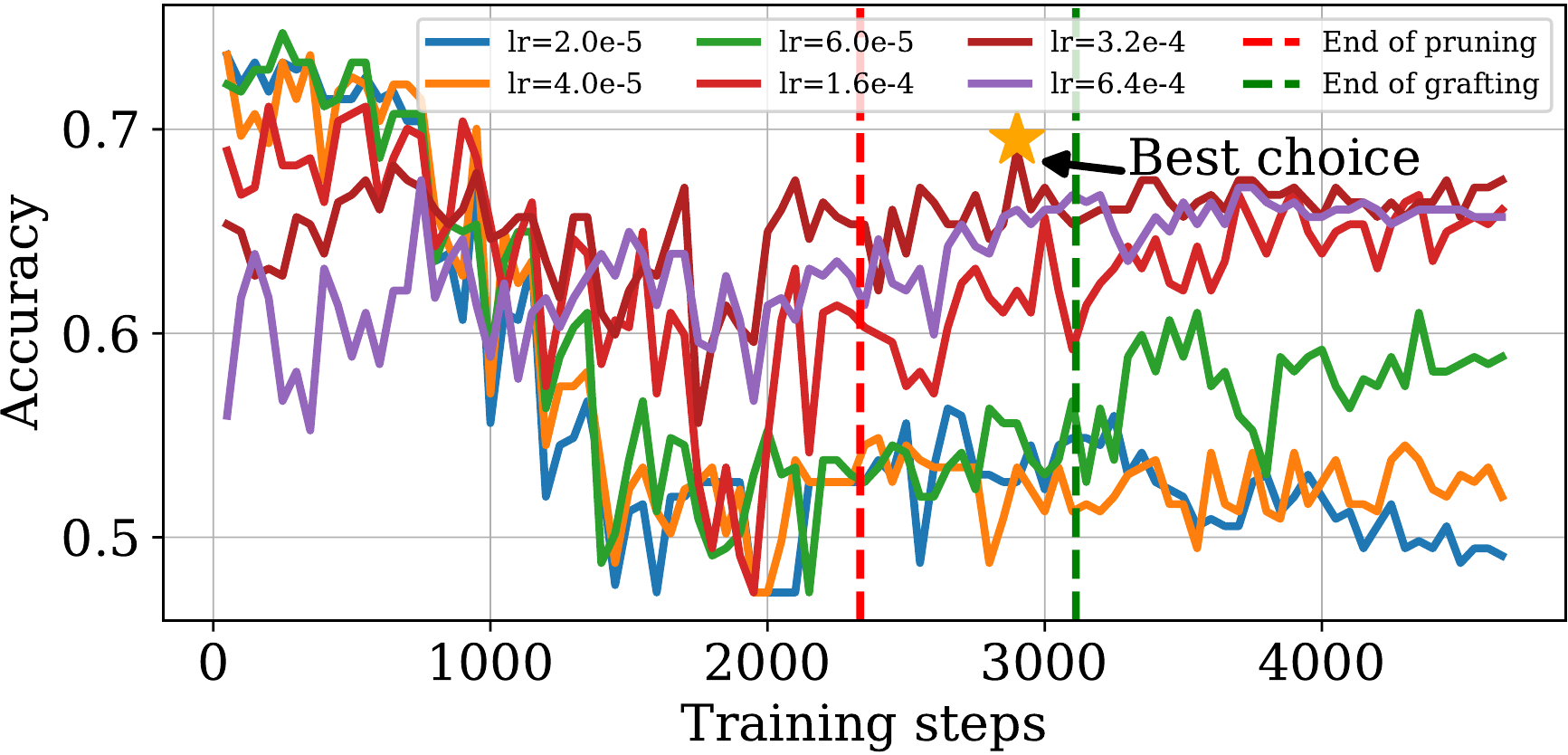}
\caption{Sensitivity analysis of learning rate on RTE (dev set).}
\label{fig:RTE_lr}
\end{figure}

\begin{figure}[b]
\centering
\includegraphics[width=1.\columnwidth,height=.210\textwidth]{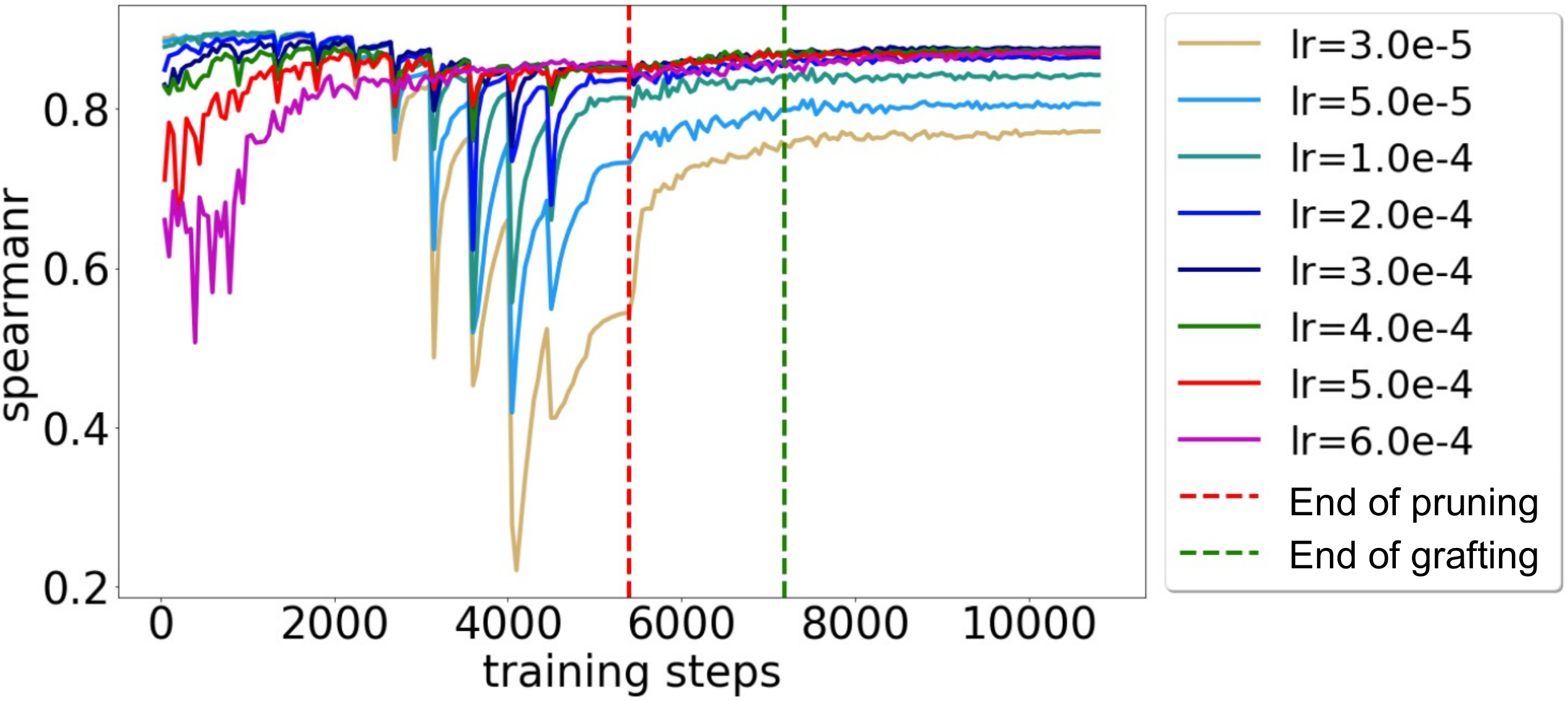}
\caption{Sensitivity analysis of learning rate on STS-B (dev set).}
\label{fig:STS-B_lr}
\end{figure}

\textbf{Sensitivity Analysis of Learning Rate.} The analysis results on RTE and STS-B are shown in Figure~\ref{fig:RTE_lr} and Figure~\ref{fig:STS-B_lr}, respectively. Results vary with different learning rate settings. Among the eight learning rates listed in the legend of Figure~\ref{fig:RTE_lr}, $3.2 \times e^{-4}$ achieves the best performance. For STS-B, $4.0 \times e^{-4}$ gives the best performance among the learning rate choices in Figures~\ref{fig:STS-B_lr}.

\begin{figure}[h]
\centering
\includegraphics[width=1.\columnwidth,height=.210\textwidth]{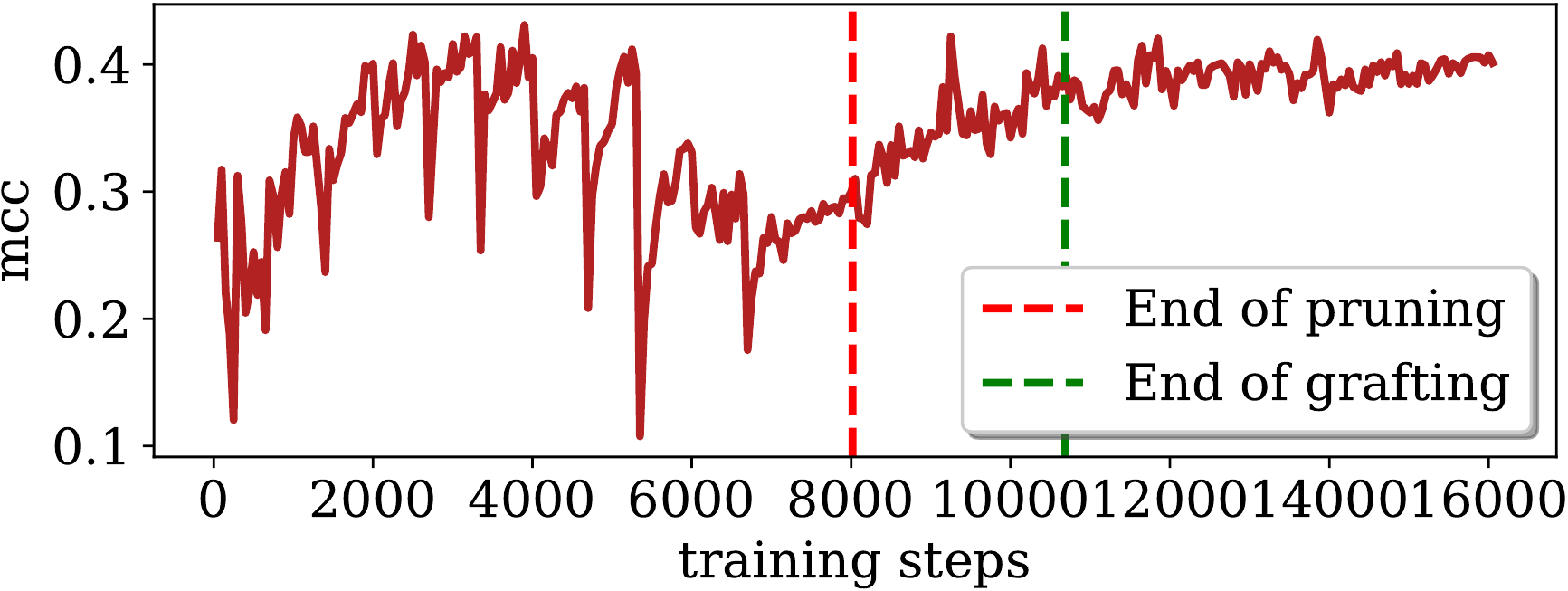}
\caption{Evaluation on CoLA (dev set). Target pruning rate is 0.95.}
\label{fig:CoLA_best}
\end{figure}

\begin{figure}[h]
\centering
\includegraphics[width=1.\columnwidth,height=.210\textwidth]{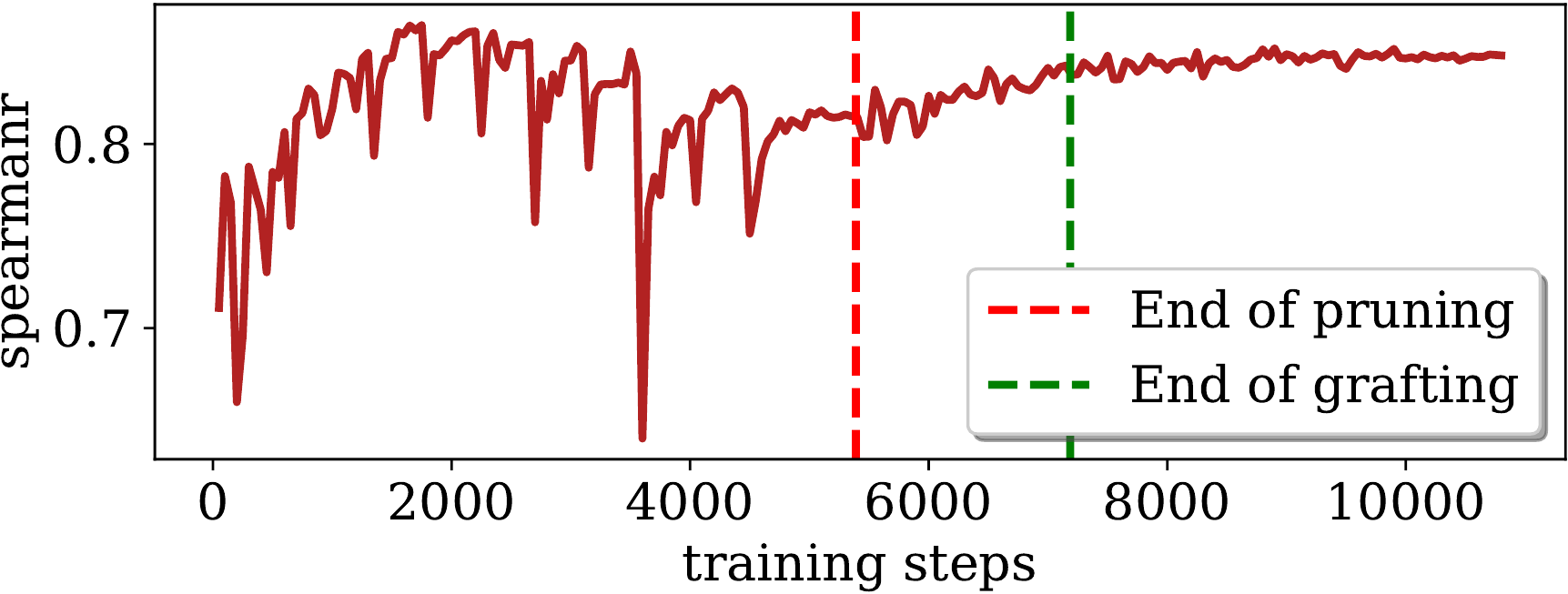}
\caption{Evaluation on STS-B (dev set). Target pruning rate is 0.95.}
\label{fig:STS-B_best}
\end{figure}

\begin{figure}[h]
\centering
\includegraphics[width=1.\columnwidth,height=.210\textwidth]{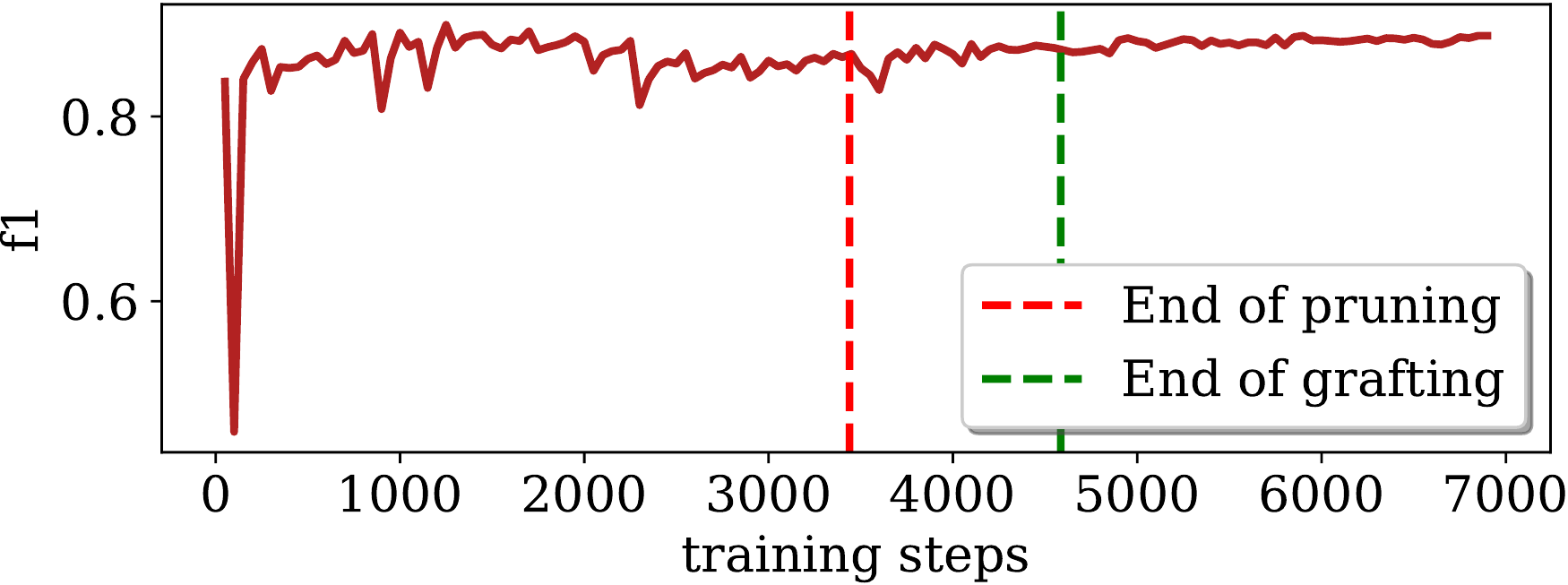}
\caption{Evaluation on MRPC (dev set). Target pruning rate is 0.95.}
\label{fig:MRPC_best}
\end{figure}

\begin{figure}[b]
\centering
\includegraphics[width=1.\columnwidth]{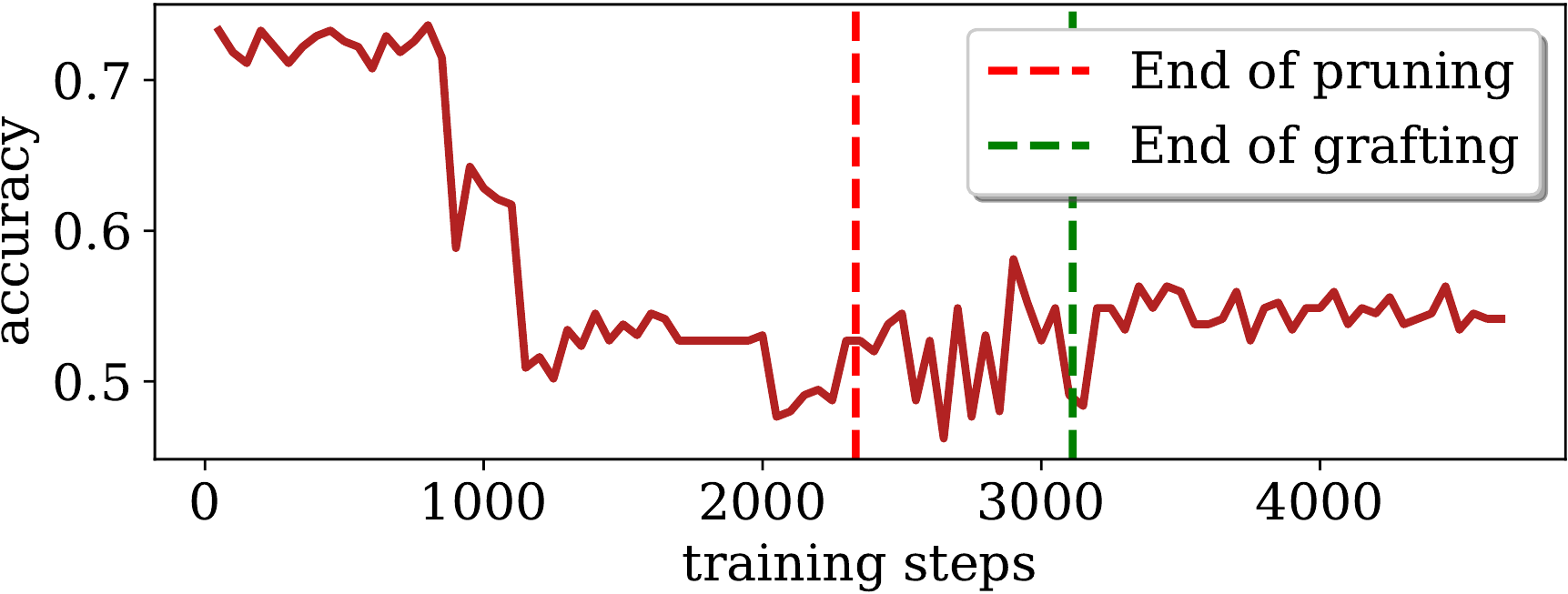}
\caption{Evaluation on RTE (dev set). Target pruning rate is 0.95.}
\label{fig:RTE_best}
\end{figure}

\textbf{Evaluation Curves of Four Tasks at Target Pruning rate of 0.95.} We plot the evaluation curves of CoLA (Figure~\ref{fig:CoLA_best}), STS-B (Figure~\ref{fig:STS-B_best}), MRPC (Figure~\ref{fig:MRPC_best}), RTE (Figure~\ref{fig:RTE_best}) 
to further demonstrate the advantages of our proposed method SPD.
In each figure, the x-axis is the training steps while the y-axis represents evaluation metrics. 
To obtain the curves, we use the same settings as 
Table~\ref{tb:compress_higher}.

Moreover, we describe the hyper-parameters settings in detail. For CoLA, we set the max sequence length as 128, the learning rate as $5.0e^{-4}$, the grafting probability during pruning as 0.8, the number of training epochs as 60, and the number of pruning epochs as 30. For STS-B, we use the same setting as CoLA. For MRPC, we set the max sequence length as 128, the learning rate as $6.4 \times e^{-4}$, the grafting probability during pruning as 0.8, the number of training epochs as 60, and the number of pruning epochs as 30. For RTE, we set the max sequence length as 128, the learning rate as $3.0 \times e^{-5}$, the grafting probability during pruning as 0.6, the number of training epochs as 60, and the number of pruning epochs as 30.

\end{document}